\pdfoutput=1
\documentclass{article}

\usepackage[preprint]{neurips_2026}

\usepackage{array}
\usepackage{tabularx}

\usepackage[utf8]{inputenc} 
\usepackage[T1]{fontenc}    
\usepackage{hyperref}       
\usepackage{url}            
\usepackage{booktabs}       
\usepackage{amsfonts}       
\usepackage{nicefrac}       
\usepackage{microtype}      
\usepackage{xcolor}         

\usepackage{amsmath, amssymb}

\usepackage{multirow}
\usepackage{graphicx}
\usepackage{enumitem}
\usepackage{caption}
\usepackage{subcaption}
\usepackage{float}
\usepackage{needspace}
\usepackage{tikz}
\usetikzlibrary{positioning,arrows.meta,calc,fit,backgrounds,shapes.multipart,shapes.geometric}
\usepackage{pgfplots}
\pgfplotsset{compat=1.18}

\newcommand{\systemName}{TeachArena}

\usepackage{makecell}
\providecommand{\eabTotalTasks}{354}

\providecommand{\eabCompleteOverlays}{17}

\definecolor{EABYes}{HTML}{1F7A3A}    
\definecolor{EABPart}{HTML}{1D4ED8}   
\definecolor{EABNo}{HTML}{B91C1C}     

\providecommand{\eabyes}{}
\providecommand{\eabpart}{}
\providecommand{\eabno}{}
\renewcommand{\eabyes}{\textcolor{EABYes}{\raisebox{0.04ex}{\scalebox{1.10}{$\checkmark$}}}}
\renewcommand{\eabpart}{\textcolor{EABPart}{\raisebox{0.12ex}{\scalebox{0.96}{$\triangle$}}}}
\renewcommand{\eabno}{\textcolor{EABNo}{\raisebox{0.02ex}{\scalebox{1.02}{$\times$}}}}

\definecolor{EABStageOne}{HTML}{2563EB}   
\definecolor{EABStageTwo}{HTML}{E98B2A}   
\definecolor{EABStageThree}{HTML}{198754} 
\definecolor{EABTagGray}{HTML}{4B5563}
\definecolor{EABRev}{HTML}{6D28D9}        
\newcommand{\rev}[1]{#1}

\providecommand{\eabstage}{}
\renewcommand{\eabstage}[2]{%
  \begingroup
  \setlength{\fboxsep}{1.15pt}%
  \raisebox{0.12ex}{%
    \colorbox{#1}{%
      \strut\hspace{0.35pt}%
      \textcolor{white}{\scriptsize\bfseries\sffamily #2}%
      \hspace{0.35pt}%
    }%
  }%
  \endgroup
}

\providecommand{\eabstagehead}{}
\renewcommand{\eabstagehead}[3]{%
  \par\addvspace{0.38em}%
  \noindent\eabstagechip{#1}{#2}\hspace{0.38em}\textbf{#3}\hspace{0.38em}%
}

\providecommand{\eabstagechip}{}
\renewcommand{\eabstagechip}[2]{\eabstage{#1}{#2}}

\providecommand{\eabtag}{}
\renewcommand{\eabtag}[1]{%
  \textcolor{EABTagGray}{\textbf{\textsc{\footnotesize #1.}}}%
  \nobreak\hspace{0.28em}\ignorespaces
}

\title{Are Agents Ready to Teach? A Multi-Stage Benchmark for Real-World Teaching Workflows}

\author{
  $\textbf{Zixin Chen}^{1,2}\thanks{Work done during internship at Qwen Team} \;\; \textbf{Peng Liu}^2 \;\; \textbf{Rui Sheng}^1 \;\; \textbf{Haobo Li}^1 \;\; \textbf{Jianhong Tu}^2$ \\ 
  $\textbf{Xiaodong Deng}^2 \;\; \textbf{Kashun Shum}^{1,2} \;\; \textbf{Dayiheng Liu}^2 \;\; \textbf{Huamin Qu}^1$ \\
  $^1$ The Hong Kong University of Science and Technology \\
  $^2$ Qwen Team, Alibaba Group \\
  \texttt{zchendf@connect.ust.hk} \\
}

\begin{document}


\maketitle

\begin{abstract}

Language agents are increasingly deployed in professional workflows, yet tutoring remains a high-stakes capability that existing evaluations only partially capture. Effective tutor agents require more than producing correct answers or executing accurate tool calls: they must infer a warranted teaching decision from evidence, adapt support as learner state changes, and carry an instructor's request through a learning-management system (LMS) to a completed, verified intervention. We introduce \textsc{\systemName{}}, a source-grounded benchmark that jointly evaluates three complementary surfaces of teaching work: professional pedagogical judgment, situated multi-turn tutoring, and end-to-end LMS teaching workflows. Its \eabTotalTasks{} tasks are each built around a pedagogical insight, grounded in evidence, and evaluated with matched verifiers over observable turn-level responses, tutoring trajectories, and persistent artifacts or environment states. Across a comprehensive evaluation of frontier models, our findings reveal that current models are generally capable of bounded pedagogical judgment, but still fall short of professional teaching standards in situated tutoring and end-to-end teaching-workflow execution. By unifying teacher judgment, adaptive tutoring, and institutional action in one benchmark, \textsc{\systemName{}} provides a measurement foundation for developing tutor agents that can support realistic teaching work.

\end{abstract}

\section{Introduction}\label{sec1}

Language agents are beginning to take on increasingly complex work across professional domains, from software engineering to customer support and even scientific discovery~\cite{ho2025verilogcoder,islam2025codesim,wang2025medkgi,chen2026vizqstudio,ghafarollahi2025sciagents,sun2026,yang2024crag,sheng2026, dong2026}. Education is a natural and consequential next frontier, but teaching places unusually broad demands on an agent~\cite{chen2026informal}. A robust tutor agent would need to do far more than answer questions: it must diagnose what a learner understands, decide what kind of support is pedagogically appropriate, sustain a productive interaction, and, in many settings, act through course-management systems where messages, grades, quizzes, and interventions affect real students~\cite{danielson2007enhancing,chen2026passive}.


Current evaluations do not yet jointly measure this full teaching object. Existing educational benchmarks often assess either subject-matter competence or local tutoring support: solving exam items, producing explanations, giving hints, or providing feedback to a single student query. These signals do not by themselves reveal whether an agent can adapt scaffolding after repeated student failure, distinguish a lucky answer from genuine understanding, or verify students' transfer of knowledge. Agent benchmarks, in contrast, increasingly test tool use and environment interaction, but rarely define success as completing a \emph{teaching workflow}. A model can call the right API, send a notification, or generate a quiz while still acting on the wrong evidence, missing the affective and motivational needs of struggling students, or creating material that fails to target the diagnosed misconception.

Grounded in pedagogical theories, we argue that tutor-agent readiness requires three separable capability surfaces. First, \textbf{pedagogical judgment} is a teacher-reasoning problem: an agent must diagnose students' misconceptions, evaluate assessment evidence, reason about prerequisites, and make intervention decisions grounded in pedagogical knowledge and formative assessment~\cite{shulman1986those,ball2008content,black1998assessment,chen2025cograder,mandinach2016data}. Second, \textbf{situated tutoring} is a trajectory-sensitive interaction problem: the same hint can be helpful early in a dialogue and harmful after a learner has already failed repeatedly, so support must adapt over time, and understanding or transfer is credited only when the learner produces the key reasoning~\cite{vygotsky1978mind,wood1976role,chen2024stugptviz,zimmerman2002self}. Third, \textbf{end-to-end LMS teaching workflow execution} is an institutional action problem: a faithful agent must translate educational decisions into evidence-grounded materials, assessments, communications, and durable changes to the right course objects. Together, these surfaces separate \emph{knowing} what a teacher should do, \emph{teaching} through an evolving interaction, and \emph{acting} without losing the educational rationale across an institutional workflow.

To provide a measurement foundation for developing future tutor agents that can support realistic teaching work, we introduce \textsc{\systemName{}}, a theory-driven and source-grounded benchmark spanning three complementary surfaces of teaching work.\footnote{All code and data are available on \href{https://huggingface.co/datasets/CinderD/TeachArena}{Hugging Face}.} \textsc{\systemName{}} contains \eabTotalTasks{} tasks across these surfaces and six teacher-work capabilities: diagnosing, designing, creating, teaching, communicating, and evaluating~\cite{danielson2007enhancing}. Rather than constructing tasks ad hoc, we use a \emph{pedagogical-insight-driven} pipeline: for each task, we specify a target instructional insight, ground it in transformed educational material, published pedagogy, or deterministic course state, and instantiate it as a teacher judgment, tutoring dialogue, or LMS workflow with observable evidence and a matched verifier. For example, one Stage~1 task uses a 120-student prerequisite pattern to test whether the agent identifies the gateway prerequisite as the defensible remediation target, rather than simply selecting the lowest-scoring downstream concept; its checks require both the decision and the mechanism that warrants it. This construction lets \textsc{\systemName{}} evaluate not only whether an agent knows the right pedagogical principle, but whether it can apply such principles through teacher judgment, adaptive dialogue, and grounded course action.

Across a comprehensive evaluation of frontier models, our findings reveal a clear gap between knowing pedagogy and enacting it. Current models often perform well on bounded pedagogical judgment, such as recognizing misconceptions, selecting feedback strategies, and answering structured teaching-decision questions. Across the \eabCompleteOverlays{}-model evaluation, however, model rankings reshuffle when judgment must become adaptive tutoring or end-to-end LMS action; our stage-resolved analysis identifies these distinct teaching-readiness profiles, while workflow scores top out at 0.704. These results motivate \textsc{\systemName{}} as a foundation for moving beyond knowing what good teaching looks like toward testing whether agents can reliably carry it out in realistic educational contexts. To summarize, our contributions are as follows:




\begin{itemize}
  \item We propose a theory-grounded framework for evaluating tutor agents across Stage~1 pedagogical judgment, Stage~2 situated tutoring, and Stage~3 end-to-end LMS teaching workflow execution, spanning six teacher-work capabilities.
  \item We introduce \textsc{\systemName{}}, a source-grounded benchmark with \eabTotalTasks{} audited tasks built through an insight-first pipeline that pairs situated educational tasks with verifiers matched to the evidence each task produces.
  \item We evaluate \eabCompleteOverlays{} frontier models and identify a knowing--teaching--acting gap through stage-rank reshuffling and localized failures in multi-turn tutoring and institutional workflow execution.
\end{itemize}

\section{Related Work}
\label{sec:related-works}

Appendix~\ref{sec:related-positioning} situates \textsc{\systemName{}}'s teaching-system episode in prior work on educational competence, tutoring, learner modeling, pedagogical decision-making, and agentic workflow evaluation.
\section{\systemName{}: Source-Grounded Benchmark Design}
\label{sec:benchmark-design}

\textsc{\systemName{}} operationalizes the \emph{teaching-system episode} through \eabTotalTasks{} auditable measurement contracts. In such an episode, evidence informs an instructional decision, the decision guides action, and the resulting learner behavior or course state can be observed.


\setlength{\intextsep}{8pt}
\begin{figure}[H]
\centering
\includegraphics[width=\textwidth]{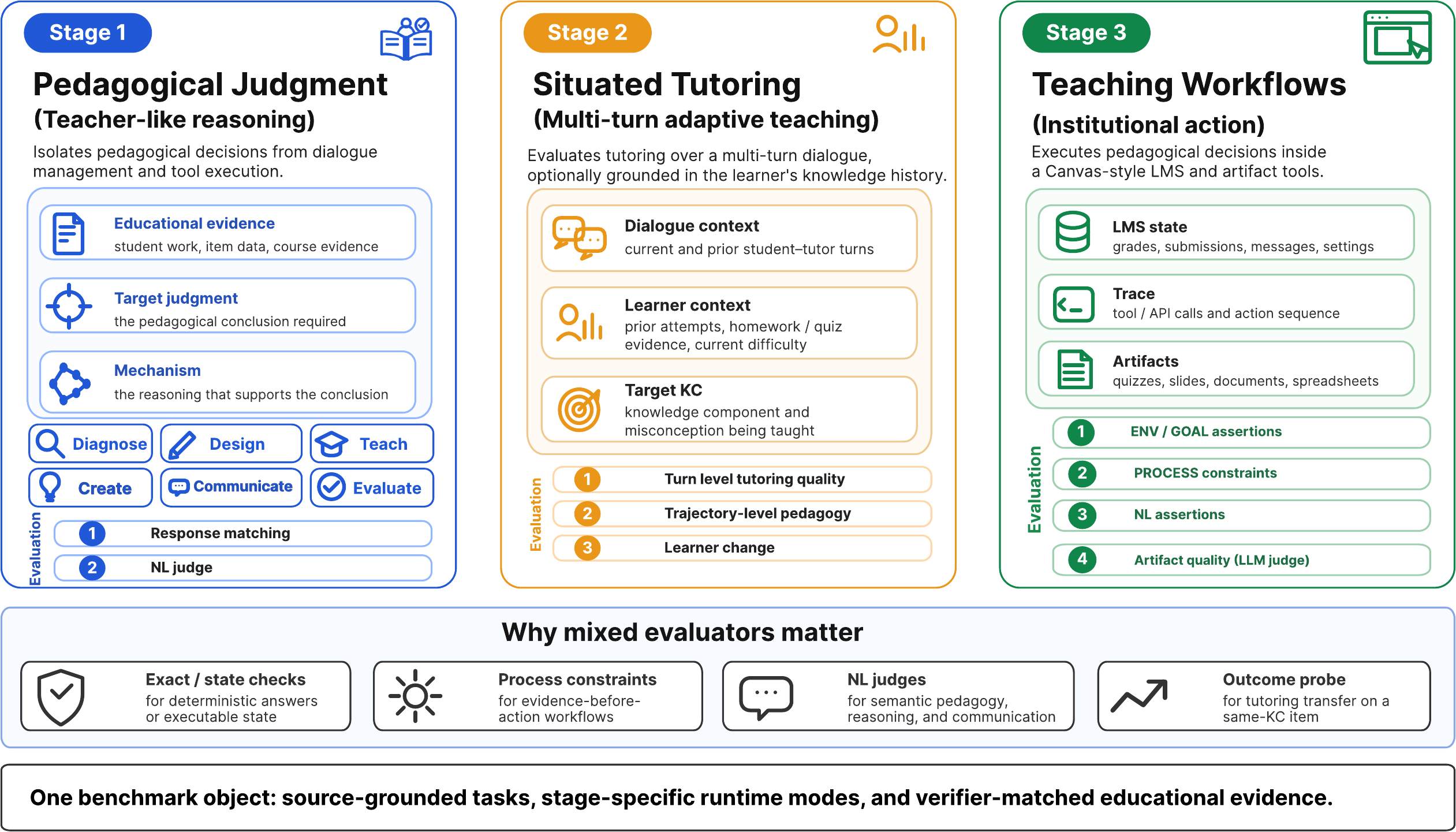}
\caption{\textbf{A decomposition of a teaching-system episode.}
Stage~1 asks for a bounded teacher decision from packaged evidence; Stage~2 tests adaptive tutoring and checks for learner-produced reasoning; Stage~3 asks it to carry an instructor's request from distributed LMS evidence to a completed, verified course intervention. Each surface leaves a different observable trace but shares one contract---educational insight, grounded evidence, observable trace, and matched verifier.}
\label{fig:method-overview}
\end{figure}
\setlength{\intextsep}{12pt plus 2pt minus 2pt}

The three stages are complementary diagnostic surfaces rather than an ordered difficulty scale. Stage~1 supplies the relevant evidence and isolates the teacher's \emph{judgment}. Stage~2 fixes the instructional target but makes each tutor move change the next learner evidence, exposing the adaptive teaching \emph{policy}. Stage~3 distributes evidence across an LMS and makes actions persistent, testing whether the agent can carry an instructor's objective from diagnosis to verified \emph{workflow} completion. This controlled decomposition operationalizes the knowing--teaching--acting story while spanning \textsc{Diagnose}, \textsc{Design}, \textsc{Create}, \textsc{Teach}, \textsc{Communicate}, and \textsc{Evaluate}~\citep{danielson2007enhancing}. Figure~\ref{fig:method-overview} summarizes the shift in evidence and observable trace across stages.

\eabstagehead{EABStageOne}{Stage 1}{Pedagogical judgment: can the agent decide what a teacher should do next?}
Stage~1 asks a bounded professional question: once the relevant learner work, assessment evidence, or instructional alternatives are on the teacher's desk, can the agent infer the warranted diagnosis or next action and explain why? The evidence is supplied, but the teaching decision is not. Freezing retrieval, dialogue, and tools reduces confounding from missing records, learner simulation, and execution. These are not subject-matter QA items: a model may recompute every number or recognize every topic and still fail by choosing a salient symptom, a generic intervention, or a fluent rationale that the evidence does not support.

The 117 tasks comprise 26 MathDial transformations~\citep{macina2023mathdial}, 17 MathTutorBench transformations~\citep{macina2025mathtutorbench}, 22 OER/literature-grounded cases, and 52 authored evidence contracts. In each task, source evidence, deterministic data, or published theory fixes the target mechanism before the decision prompt and matched verifier are written; Appendix~\ref{app:stage1-design-audit} details provenance and construction.

\begin{table}[H]
\caption{\textbf{Stage~1 task taxonomy.} Each family is defined by the teaching decision that is ultimately scored.}
\label{tab:stage1-decision-spans}
\centering
\footnotesize
\setlength{\tabcolsep}{2.5pt}
\renewcommand{\arraystretch}{0.96}
\begin{tabularx}{\columnwidth}{@{}>{\raggedright\arraybackslash}p{0.23\columnwidth}>{\raggedright\arraybackslash}p{0.18\columnwidth}X@{}}
\toprule
\textbf{Task family} & \textbf{Decision object} & \textbf{Scored judgment} \\
\midrule
\textbf{Learner-state inference} & Learner or cohort state & \textbf{Explain} the evidence by identifying the latent misconception, prerequisite gap, or cognitive, metacognitive, or affective state~\citep{shulman1986those,ball2008content}. \\
\textbf{Instructional-action selection} & Teaching move & \textbf{Choose or critique} an explanation, representation, feedback strategy, scaffold, placement, or material for the learner and moment~\citep{black1998assessment,wood1976role}. \\
\textbf{Assessment and claim audit} & Assessment artifact or empirical claim & \textbf{Audit} whether an item, score interpretation, or data claim warrants its conclusion; state the needed revision or qualification~\citep{stiggins2002assessment,mandinach2016data}. \\
\bottomrule
\end{tabularx}
\end{table}

\eabtag{Illustrative case: same score, opposite actions} \texttt{EI-AFFECT-01} asks the model to recommend the next teaching move for four learners who complete the same algebra session and all score 3/10. Consider Leo and Tom. Leo slows down across the session, fails the same problem five times, and does not revise after opening a hint; he needs guided help to exit the failure loop. Tom is working above his current level, uses hints independently, self-corrects, and improves from 0/10 to 3/10 on the same items; lowering the difficulty would interrupt productive progress. The identical score therefore warrants opposite actions---intervene for Leo, but maintain the challenge for Tom. A score-only policy cannot make this distinction; the task tests whether the agent can map process evidence to a defensible teaching action. Appendix~\ref{app:stage1-discriminating} gives the complete four-learner contract, theoretical grounding, and verifier.

\eabstagehead{EABStageTwo}{Stage 2}{Situated tutoring: can the agent sustain an adaptive tutoring policy across a multi-turn episode?}
Stage~2 asks whether an agent can teach through a complete multi-turn episode, rather than produce one good tutor response. Every task controls the learner's starting point---the target concept, prior knowledge, misconception, and metacognitive and non-cognitive state---while leaving the teaching path open. The agent must use each learner reply as new evidence: elicit reasoning, update its diagnosis, adapt or fade support, and ultimately elicit an explanation or transfer from the learner. The object of evaluation is therefore a history-dependent teaching policy and learner-produced evidence observed within the episode, not isolated helpfulness or content correctness.

We organize measurement along two complementary axes. The \emph{instructional axis} covers cognitive, metacognitive, and non-cognitive teaching---for example, diagnosing a misconception, prompting the learner to check a strategy, or restoring engagement---with the corresponding learner-state controls summarized in Table~\ref{tab:stage2-learner-design}. The \emph{evidence axis} separates the local quality of each tutor turn, the adaptation and completeness of the full teaching trajectory, and learner-produced change (Table~\ref{tab:stage2-judging-dimensions}). Together, the two axes prevent one correct explanation from standing in for effective multi-turn teaching.

\begingroup
\setlength{\intextsep}{5pt}
\begin{table}[H]
\captionsetup{skip=3pt}
\caption{\textbf{Stage~2 controlled learner design.} State dimensions (top) define the simulator contract and what may change under tutoring; six authored profiles (bottom) cross them into distinct adaptation pressures. Profiles are test conditions, not learner diagnoses, and a separately prompted fidelity judge enforces the assigned state.}
\label{tab:stage2-learner-design}
\centering
\footnotesize
\setlength{\tabcolsep}{2.4pt}
\renewcommand{\arraystretch}{0.92}
\begin{tabularx}{\textwidth}{@{}>{\raggedright\arraybackslash}p{1.65cm}>{\raggedright\arraybackslash}p{4.25cm}X@{}}
\multicolumn{3}{@{}l}{\textbf{A. Learner-state dimensions}} \\[-1pt]
\toprule
\textbf{Learner dimension} & \textbf{Controlled state} & \textbf{Why it changes tutor policy} \\
\midrule
\textbf{Cognitive} & Prior mastery; known/unknown KCs; misconception; predicted errors; explanation-load tolerance. & Bounds what to diagnose, represent, or scaffold~\citep{corbett1994knowledge,sweller1988cognitive}. \\
\textbf{Metacognitive} & Planning; strategy choice; self-monitoring; reflection. & Determines when to prompt checking, fade support, and return agency~\citep{zimmerman2002self}. \\
\textbf{Non-cognitive} & Motivation; persistence; affect; coarse behavioral cues. & Determines when to challenge, reassure, or re-engage~\citep{ryan2000intrinsic,pekrun2006control}. \\
\bottomrule
\end{tabularx}
\vspace{2pt}
\begin{tabularx}{\textwidth}{@{}>{\raggedright\arraybackslash}X>{\raggedright\arraybackslash}X>{\raggedright\arraybackslash}X@{}}
\multicolumn{3}{@{}l}{\textbf{B. Six controlled profile composites}} \\[-1pt]
\toprule
\textbf{Brittle expert}\newline Fluent; one conceptual gap. & \textbf{Motivated novice}\newline Curious; limited prior knowledge. & \textbf{Deep learner}\newline Self-regulated; mechanism-seeking. \\
\textbf{Strategic surface learner}\newline Shortcut-seeking. & \textbf{Anxious achiever}\newline Prepared; error-sensitive. & \textbf{Disengaged/amotivated}\newline Low persistence; low perceived value. \\
\bottomrule
\end{tabularx}
\end{table}
\endgroup

Stage~2 contains 100 tasks across 13 disciplines: 64 new OER/MathDial-grounded scenarios and 36 adaptations of MATH, MMLU-Pro, MIT OpenCourseWare, or topic-matched hard items~\citep{macina2023mathdial,hendrycks2021math,wang2024mmlupro,mitocw}. For each task, we verify the source solution and target skill, author a learner profile whose prior knowledge and misconception predict concrete errors, and reserve a non-identical problem requiring the same skill for transfer.

Learner replies are generated with Qwen3-Max (\(T=0\)) conditioned on the hidden profile and dialogue history. Role-separated fidelity and learner-state calls use distinct prompts and structured outputs: the fidelity check rejects replies that contradict the assigned knowledge, misconception, affect, or prior turns, and the state evaluator records change in reasoning, self-monitoring, and engagement after each accepted reply. Only accepted replies are shown to the tutor, which sees neither the profile nor evaluator outputs~\citep{wu2025imperfection,chi2014icap}.

Table~\ref{tab:stage2-judging-dimensions} assigns each evaluator a distinct unit of evidence. Its learner-change scope is revelation-aware: it compares each profile-consistent learner reply with the learner's prior state and the preceding tutor turn, crediting only explanation, application, self-checking, or engagement that the learner newly produces. Agreement or repetition after the tutor supplies the key idea receives no learner-change credit. Appendix~\ref{app:stage2-design-audit} gives the complete construction, runtime controls, scoring realization, and learner-change reporting.

\begingroup
\setlength{\intextsep}{5pt}
\begin{table}[H]
\captionsetup{skip=3pt}
\caption{\textbf{Stage~2 evidence scopes.} Local tutor quality, the multi-turn teaching policy, and learner-produced change require different evidence. Cross-turn adaptation and completion of the teaching arc are two aspects of the trajectory scope.}
\label{tab:stage2-judging-dimensions}
\centering
\footnotesize
\setlength{\tabcolsep}{2.4pt}
\renewcommand{\arraystretch}{0.92}
\begin{tabularx}{\textwidth}{@{}>{\raggedright\arraybackslash}p{1.55cm}>{\raggedright\arraybackslash}p{4.10cm}X@{}}
\toprule
\textbf{Scope} & \textbf{Evidence read} & \textbf{Question and checks} \\
\midrule
\textbf{Tutor turn} & One tutor response, the immediately preceding learner turn, and recent context. & \emph{Was this move locally sound?} Identifies the teaching move and scores correctness, relevance, contingency, clarity/load, mistake localization, immediate affective response, elicited uptake, and information revelation~\citep{black1998assessment,wood1976role}. \\
\textbf{Tutor trajectory} & The complete dialogue plus the sequence of turn annotations (e.g., strategy, support, revelation, and learner state). & \emph{Did the moves compose into an effective teaching policy?} Checks strategy changes after new evidence, escalation and fading, representation pivots, productive struggle, misconception repair, agency handoff, and verification or transfer before closure~\citep{vandepol2010scaffolding,vosniadou1992mental,chi2014icap}. \\
\textbf{Learner change} & Each reply that passes the profile-consistency check, the learner's state before the exchange, and the preceding tutor turn. & \emph{What can the learner now do without being given the answer?} Credits independently produced explanation, application, self-checking, engagement, or transfer; discounts agreement or repetition of reasoning the tutor just supplied~\citep{chi2014icap,zimmerman2002self}. \\
\bottomrule
\end{tabularx}
\end{table}
\endgroup

We also log an auxiliary same-KC transfer probe, outside the headline score: a fixed probe model that fails the cold item receives the tutoring trajectory before attempting a non-identical item targeting the same KC (Appendix~\ref{app:stage2-design-audit}).

These evidence scopes can disagree even when the learner reaches the correct answer: Section~\ref{sec:diagnostics-cases} traces one episode in which local tutor quality, the multi-turn policy, and learner-owned change yield different diagnoses.

\eabstagehead{EABStageThree}{Stage 3}{End-to-end LMS workflows: can the agent carry a teacher's request to closure?}
\label{sec:stage3}
Stage~3 asks whether an agent can serve as a bounded operational delegate for an instructor inside an LMS, rather than merely recommend an action or produce a plausible file. A request may name a course, evidence source, or object to change; the agent must still recover the governing records, derive the warranted teaching intervention, execute it in the designated objects, communicate to the authorized audience, and leave the promised course state complete. The evaluated object is continuity across this evidence--decision--action chain. A trajectory can therefore look productive yet fail if one handoff changes its educational meaning---for example, by using an irrelevant assessment, placing correct content in a new deck instead of the assigned one, or announcing work before it exists.

The 137 tasks run in a deterministic, synthetic Canvas-style LMS linking 126 learners, 432 course enrollments, 222 assignment records, 1,058 quiz attempts, and 13 editable presentations to artifact and analysis tools; no record belongs to a real learner. Tasks span evidence-conditioned assessment and remediation, diagnosis-linked grading and feedback, privacy-sensitive outreach, construct-aligned material revision, and multi-artifact course updates. We author each task as an initial-state/target-state contract: seed the records that warrant a teaching decision; designate the objects and recipients through which it must be enacted; specify essential evidence-before-action dependencies and a durable completion condition; and include at least one executable but educationally invalid route. This operationalizes data-informed teaching, assessment literacy, instructional design, and constructive alignment: the evidence must warrant the intervention, and its material, assessment, and communication must remain mutually consistent~\citep{stiggins2002assessment,mandinach2016data,branch2009instructional,biggs1996enhancing}. A reference trajectory and construction-time oracle establish that the declared conditions are jointly satisfiable without prescribing the model's route.

Verification mirrors the task contract (Table~\ref{tab:stage3-measurement-contract}). Environment and goal-state checks establish what persisted and in which object; process checks establish whether prerequisite evidence preceded consequential action; and task-specific semantic assertions test whether the educational rationale survived across artifacts and audiences. These views are complementary: a well-written artifact cannot repair a wrong source, object, recipient, or dependency order.

For the 83 tasks that create or revise instructional materials, the evaluator also reads the persisted content in the context of the task---not merely a file-exists flag or tool metadata. Quiz rubrics combine multiple-choice item quality~\citep{haladyna1993}, alignment to the diagnosed KCs~\citep{biggs1996enhancing}, learner-calibrated difficulty~\citep{vygotsky1978mind}, and cognitive-level diversity~\citep{krathwohl2002revision}. Slide rubrics combine alignment, worked-example and cognitive-load principles, and an opportunity for active application~\citep{biggs1996enhancing,sweller1988cognitive,chi2014icap}. Document, LMS-page, and spreadsheet rubrics test accuracy, task fulfillment, clarity, organization, and audience usability. Artifact-quality scoring is type-specific; in a workflow with several artifacts, companion outputs not scored by that rubric remain covered by state and semantic checks.

\begingroup
\setlength{\intextsep}{3pt}
\begin{table}[H]
\captionsetup{skip=2pt}
\caption{\textbf{Stage~3 measurement contract.} Workflow checks establish that the right intervention was executed; artifact rubrics evaluate the resulting instructional content.}
\label{tab:stage3-measurement-contract}
\centering
\footnotesize
\setlength{\tabcolsep}{3pt}
\renewcommand{\arraystretch}{0.94}
\begin{tabularx}{\textwidth}{@{}>{\raggedright\arraybackslash}p{2.15cm}>{\raggedright\arraybackslash}X@{}}
\toprule
\textbf{Layer} & \textbf{Criteria} \\
\midrule
\multicolumn{2}{@{}l}{\textit{Workflow validity}} \\
\addlinespace[1pt]
\textbf{State}\newline{\scriptsize Environment + goal} &
The requested change persists in the designated object, with the required content and authorized recipients. \\
\textbf{Trace}\newline{\scriptsize Process} &
Task-specific dependencies hold (e.g., retrieve before compute, inspect before edit, create before announce). \\
\textbf{Alignment}\newline{\scriptsize Semantic} &
Evidence, diagnosis, artifact, assessment, and communication express one consistent intervention, with appropriate privacy and audience. \\
\addlinespace[3pt]
\multicolumn{2}{@{}l}{\textit{Artifact quality (83 tasks)}} \\
\addlinespace[1pt]
\textbf{Quiz} &
Valid key and wording; KC/construct alignment; diagnostic distractors; calibrated difficulty; cognitive-level diversity~\citep{haladyna1993,biggs1996enhancing,vygotsky1978mind,krathwohl2002revision}. \\
\textbf{Slides} &
Conceptual clarity and worked example; KC targeting; managed cognitive load; opportunity for active application~\citep{biggs1996enhancing,sweller1988cognitive,chi2014icap}. \\
\textbf{Other artifacts} &
Documents, LMS pages, and spreadsheets: accuracy; task-contract coverage; organization; clarity; audience usability. \\
\bottomrule
\end{tabularx}
\end{table}
\endgroup

Together, these checks define workflow success as preserving one instructional objective across evidence, intervention, designated objects, authorized audiences, and persistent course state. Section~\ref{sec:diagnostics-cases} uses a complete workflow to show how matched checks localize distinct breaks in this chain; Appendix~\ref{app:stage3-design-audit} gives five additional workflow contracts.

This diagnostic decomposition makes the knowing--teaching--acting distinction operational. Stage~1 measures bounded judgment, Stage~2 measures policy under evolving learner evidence, and Stage~3 measures continuity between evidence, intervention, and verified LMS state. Together, the stage scores form a capability profile that distinguishes relative strengths across these teaching surfaces.

\section{\rev{Results}}
\label{sec:results}

\begin{table}[!t]
\centering
\scriptsize
\setlength{\tabcolsep}{2.6pt}
\renewcommand{\arraystretch}{0.98}
\newcommand{\eabstageonetop}[1]{\begingroup\setlength{\fboxsep}{1.1pt}\colorbox{EABStageOne!12}{\textcolor{EABStageOne}{\textbf{#1}}}\endgroup}
\newcommand{\eabstagetwotop}[1]{\begingroup\setlength{\fboxsep}{1.1pt}\colorbox{EABStageTwo!14}{\textcolor{EABStageTwo}{\textbf{#1}}}\endgroup}
\newcommand{\eabstagethreetop}[1]{\begingroup\setlength{\fboxsep}{1.1pt}\colorbox{EABStageThree!12}{\textcolor{EABStageThree}{\textbf{#1}}}\endgroup}
\begin{tabular}{clcccc}
\toprule
\textbf{\#} & \textbf{Model} & \textbf{Stage 1} & \textbf{Stage 2} & \textbf{Stage 3} & \textbf{Overall} \\
\midrule
1 & Claude Opus 4.8 & \eabstageonetop{0.930} & \eabstagetwotop{0.773} & \eabstagethreetop{0.704} & \textbf{0.803} \\
2 & Claude Opus 4.6 & \eabstageonetop{0.947} & 0.763 & \eabstagethreetop{0.661} & 0.791 \\
3 & GPT-5.5-pro & 0.916 & 0.739 & \eabstagethreetop{0.693} & 0.783 \\
4 & GPT-5.5 & 0.916 & 0.717 & 0.659 & 0.764 \\
5 & Qwen3.7-Max & 0.914 & \eabstagetwotop{0.766} & 0.580 & 0.753 \\
6 & GLM-5.2 & \eabstageonetop{0.932} & 0.763 & 0.562 & 0.752 \\
7 & Qwen3.6-Plus & 0.905 & 0.756 & 0.585 & 0.749 \\
8 & GPT-5.4 & 0.874 & 0.729 & 0.596 & 0.733 \\
9 & DeepSeek-V4-Pro & 0.885 & 0.738 & 0.572 & 0.731 \\
10 & GLM-5.1 & 0.871 & 0.762 & 0.547 & 0.727 \\
11 & Gemini-3.1-pro & 0.892 & 0.763 & 0.514 & 0.723 \\
12 & Gemini-2.5-pro & 0.867 & \eabstagetwotop{0.769} & 0.477 & 0.704 \\
13 & Gemini-3.5-flash & 0.906 & 0.737 & 0.432 & 0.692 \\
14 & Kimi K2.6 & 0.901 & 0.761 & 0.395 & 0.686 \\
15 & Qwen3-Max & 0.900 & 0.683 & 0.449 & 0.677 \\
16 & Gemini-3.1-flash-lite & 0.819 & 0.748 & 0.452 & 0.673 \\
17 & GPT-4.1 & 0.799 & 0.732 & 0.447 & 0.659 \\
\bottomrule
\end{tabular}
\caption{\textbf{Leaderboard across the three benchmark stages.} Overall is the unweighted mean of the three stage scores. Rows follow Overall; blue, orange, and green cells mark each stage's top three.}
\label{tab:main-results-v29}
\end{table}

\paragraph{\rev{Stage profiles reveal complementary capability bottlenecks.}}
\rev{Table~\ref{tab:main-results-v29} reports performance for 17 models. Stage~1 scores are high under its bounded-evidence contract (mean 0.893; range 0.799--0.947), while stage-resolved reporting reveals substantial rank variation. The descriptive rank correlations are weak between Stage~2 and both Stage~1 (\(\rho=0.24\)) and Stage~3 (\(\rho=0.21\)), whereas Stages~1 and~3 are more closely associated (\(\rho=0.58\)); the median difference between a model's best and worst stage rank is seven positions. The benchmark decomposition makes these shifts interpretable. Stage~1 isolates a bounded teacher decision under packaged evidence. Stage~2 separately measures whether a history-dependent teaching policy adapts to learner responses while preserving learner agency; knowing an appropriate move in isolation does not determine which systems can sustain an adaptive tutoring policy. Stage~3 requires educational diagnosis to compose with retrieval, planning, reliable tool execution, persistent state or artifact changes, communication, and verification. Because this contract is conjunctive, a weak handoff can prevent verified completion even when other components are strong. The resulting profiles distinguish relative strengths in bounded judgment, adaptive tutoring, and end-to-end execution.}

\paragraph{\rev{Frontier systems remain uneven across teaching capabilities.}}
\rev{Claude Opus~4.8 has the highest overall point estimate (0.803) and is the only model whose point estimates rank in the top three on all three stages. Even this broadly strong profile does not saturate the matched situated-tutoring or workflow contracts, scoring 0.773 and 0.704, respectively. All four Gemini variants rank lower on Stage~3 than on Stage~2; Gemini-2.5-Pro is second on tutoring (0.769) but 12th on workflows (0.477). This pattern echoes the education-focused behavior targeted by LearnLM~\citep{learnlm2024}, although proprietary training provenance precludes causal attribution. Conversely, GPT-5.5 ranks fourth on workflows but 16th on tutoring. Stage-resolved reporting further differentiates systems with similar overall performance: Kimi~K2.6 and Qwen3-Max differ by only 0.009 overall, although Kimi is 0.078 higher on tutoring and 0.054 lower on workflows. Together, the overall and stage-resolved views summarize broad competence while identifying whether improvement should target adaptive teaching policy or reliable end-to-end execution.}

\paragraph{\rev{Two stress conditions recur across systems.}}
\rev{In Stage~2, the strategic surface learner is the lowest-scoring of six profiles for 15 of 17 models, averaging 0.702 versus 0.757 across the other profiles. Unlike a disengaged learner, this profile remains interactive but solicits shortcuts or answer confirmation; the gap is consistent with difficulty preserving learner agency under apparent cooperation. In Stage~3, the six long-chain workflows average 0.321 versus 0.559 across all other Stage~3 tasks and are the lowest-scoring family for 15 of 17 models. This association is consistent with composition pressure across evidence, actions, artifacts, persistent state, and communication, without identifying chain length or any interface as the cause. Together, these descriptive contrasts identify recurring stress conditions in maintaining productive learner ownership and preserving an educational objective across a multi-interface workflow. Complete breakdowns, model-level exceptions, bootstrap confidence intervals, and the analysis protocol appear in Appendix~\ref{app:diagnostic-results}.}

\section{\rev{Diagnostics and Case Studies}}
\label{sec:diagnostics}
\label{sec:diagnostics-cases}

\rev{Aggregate rewards show that a contract is incomplete but not which dependency failed. We therefore audit two final-release cases with matched evidence scopes: the Stage~2 episode with historical release ID \texttt{S0-54} and the Stage~3 workflow \texttt{MM-04}. They illustrate observable failure modes, not their prevalence.}

\paragraph{\rev{Stage 2: a correct answer can still leave the teaching arc incomplete.}}
\rev{In \texttt{S0-54}, a first-year statistics learner asks whether a positive medical test means that she has a \(99\%\) chance of disease. The disease prevalence is \(1\%\); the test has \(99\%\) sensitivity but also returns positive for \(5\%\) of people without the disease. Her answer, \(0.99\), reverses the conditional---it substitutes \(P(+\mid D)\) for \(P(D\mid+)\)---and ignores prevalence. Among 10,000 people, the expected counts are 99 true positives and 495 false positives, so \(P(D\mid+)=99/(99+495)=1/6\). The learner is also anxious and may abandon valid reasoning when challenged, so effective tutoring must repair the cognitive error without turning reassurance into confirmation. In the evaluated trace, natural frequencies help the learner compute \(1/6\) and explain why the base rate matters~\citep{gigerenzer1995}. Table~\ref{tab:stage2-case-judge-trace} shows why this correct outcome still receives a mixed diagnosis.}

\begingroup
\setlength{\intextsep}{5pt}
\begin{table}[H]
\captionsetup{skip=3pt}
\caption{\textbf{What the three Stage~2 evidence scopes reveal in the same \texttt{S0-54} trace.} Although the learner reaches the correct posterior, local tutor quality, the multi-turn policy, and learner-owned change yield different diagnoses.}
\label{tab:stage2-case-judge-trace}
\centering
\scriptsize
\setlength{\tabcolsep}{2.5pt}
\renewcommand{\arraystretch}{0.96}
\begin{tabularx}{\textwidth}{@{}>{\raggedright\arraybackslash}p{2.35cm}>{\raggedright\arraybackslash}p{5.10cm}X@{}}
\toprule
\textbf{Scope and question} & \textbf{Evidence in the \texttt{S0-54} trace} & \textbf{Case-specific finding} \\
\midrule
\textbf{Tutor turn}\newline\emph{Was this move appropriate now?}
& \emph{Opening:} responding to the learner's anxious request for confirmation, the tutor first steadies her, then distinguishes the two conditional directions and asks her to count true and false positives among 100 people. \emph{Later:} after she computes \(0.167\), the tutor explains that this means most positive results are false positives.
& The opening is credited for accurate diagnosis, affect-first support, and preserved reasoning demand. The later turn is downgraded: its content is correct, but it supplies the interpretation the learner should construct. \\
\textbf{Tutor trajectory}\newline\emph{Did the moves form an effective policy?}
& Across replies, the tutor moves from reassurance to a natural-frequency decomposition and then fades to verification after the learner computes \(0.167\). End to end, the learner explains why the base rate matters, but the tutor supplies the interpretation that most positives are false and no cross-context transfer item follows.
& Credits state-contingent strategy and support changes, followed by conceptual repair. Full trajectory credit is withheld because part of the interpretation remains tutor-owned, the agency handoff is incomplete, and the dialogue ends before any cross-context transfer item is attempted. \\
\textbf{Learner change}\newline\emph{What did the learner demonstrate independently?}
& The learner opens with ``Please tell me I did that right.'' She later derives \(1/6\), reapplies the method in the same medical-test scenario with prevalence set to \(50\%\), and explains: ``the base rate really matters~\ldots{} I totally ignored how rare the disease was.''
& Credits the independent calculation, changed-prevalence application, and shift from anxious confirmation-seeking to explanatory engagement; later reuse of tutor-supplied language receives discounted credit. \\
\bottomrule
\end{tabularx}
\end{table}
\endgroup

\rev{The findings are complementary: within the simulated trace, the learner repairs the inverse-conditional misconception and remains engaged, but a later tutor move supplies an interpretation the learner should construct, leaving the agency handoff partial; the learner reapplies the method at a different prevalence in the same medical-test scenario. The dialogue ends before any cross-context transfer item is attempted. An answer-only check would record success, whereas the matched scopes distinguish learner-owned repair from tutor-owned explanation and missing verification.}

\paragraph{\rev{Stage 3: educational validity must survive every system boundary.}}
\rev{In \texttt{MM-04}, an ECON101 instructor requests visual supply--demand remediation using the Fall~2023 midterm and the existing Week~5 deck. The environment also contains an easier-to-find current quiz, but it tests opportunity cost, comparative advantage, and production possibilities and is therefore an invalid source. From 120 historical attempts, the agent must recover four weak KCs---elasticity and tax incidence (\(13/120\) correct each), consumer surplus (\(20/120\)), and equilibrium (\(30/120\))---then inspect and revise at least two slides in the named deck, add at least three relevant economics diagrams, create a three-question visual quiz, and only then notify the advisor and class using aggregate evidence. The case binds one diagnosis to a specific instructional artifact, assessment, audience, and completion order; using the distractor quiz, creating a detached deck, or announcing unfinished work each preserves some local competence while breaking the delegated teaching objective.}

\begin{figure}[t]
\centering
\includegraphics[width=\textwidth]{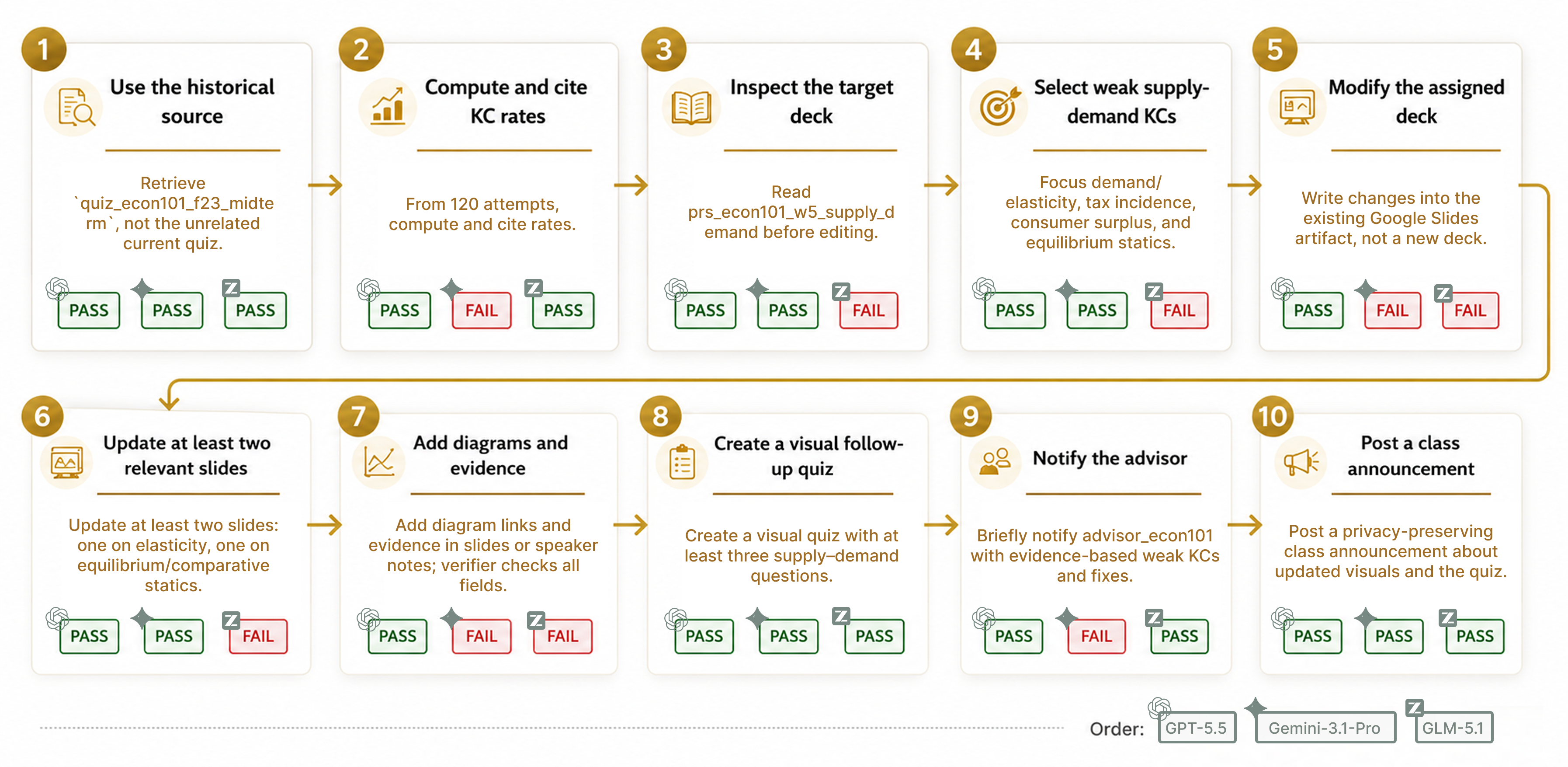}
\caption{\textbf{Verifier-level trajectory for \texttt{MM-04}.} The task requires an evidence-to-action chain: retrieve the correct historical quiz, compute KC weaknesses, inspect the assigned deck, edit the existing teaching artifact, create a targeted quiz, and communicate the intervention. Step-level pass/fail markers show where representative model trajectories satisfy or miss the teaching-work contract.}
\label{fig:case-artifacts-mm04-v6}
\end{figure}

\rev{The checks make every declared boundary in Figure~\ref{fig:case-artifacts-mm04-v6} auditable. Their checkpoints encode educational obligations rather than backend trivia. State predicates require the \emph{existing} Week~5 deck---not just any file---to contain the required slide and diagram changes, and require the quiz and both communications to persist. Process constraints require querying historical submissions before computation, deck inspection before editing, computation before material design, and completed materials before announcement. Semantic assertions bind the source, rates, diagram explanations, quiz target, and privacy-preserving messages to the same four KCs. Finally, \texttt{MM-04}'s declared slide rubric examines the persisted revisions for conceptual clarity, worked examples, KC targeting, cognitive load, and engagement; the companion quiz is checked separately for existence and supply--demand alignment. A reference execution satisfies the full stack, while the representative overlays in the figure fail at different handoffs. This decomposition also explains partial progress: a trajectory can retrieve the correct record or send a plausible message while the evidence never reaches the assigned artifact.}

\rev{The three selected trajectories exhibit distinct outcomes. GPT-5.5 closes the loop: diagnosed weaknesses become visual slide remediation, the quiz targets the same concepts, and the messages refer to that completed intervention. GLM-5.1 crosses the content boundary but not the object-identity boundary: it produces plausible microeconomics material while leaving the assigned deck unchanged and drifting toward generic practice. Gemini-3.1-Pro reaches the designated artifact but misses required numeric or visual evidence. Similar-looking activity can therefore have different evidential validity. These are not three degrees of prose quality, but failures at different links in one teaching contract.}

\begin{figure}[!h]
\centering
\includegraphics[width=0.96\linewidth]{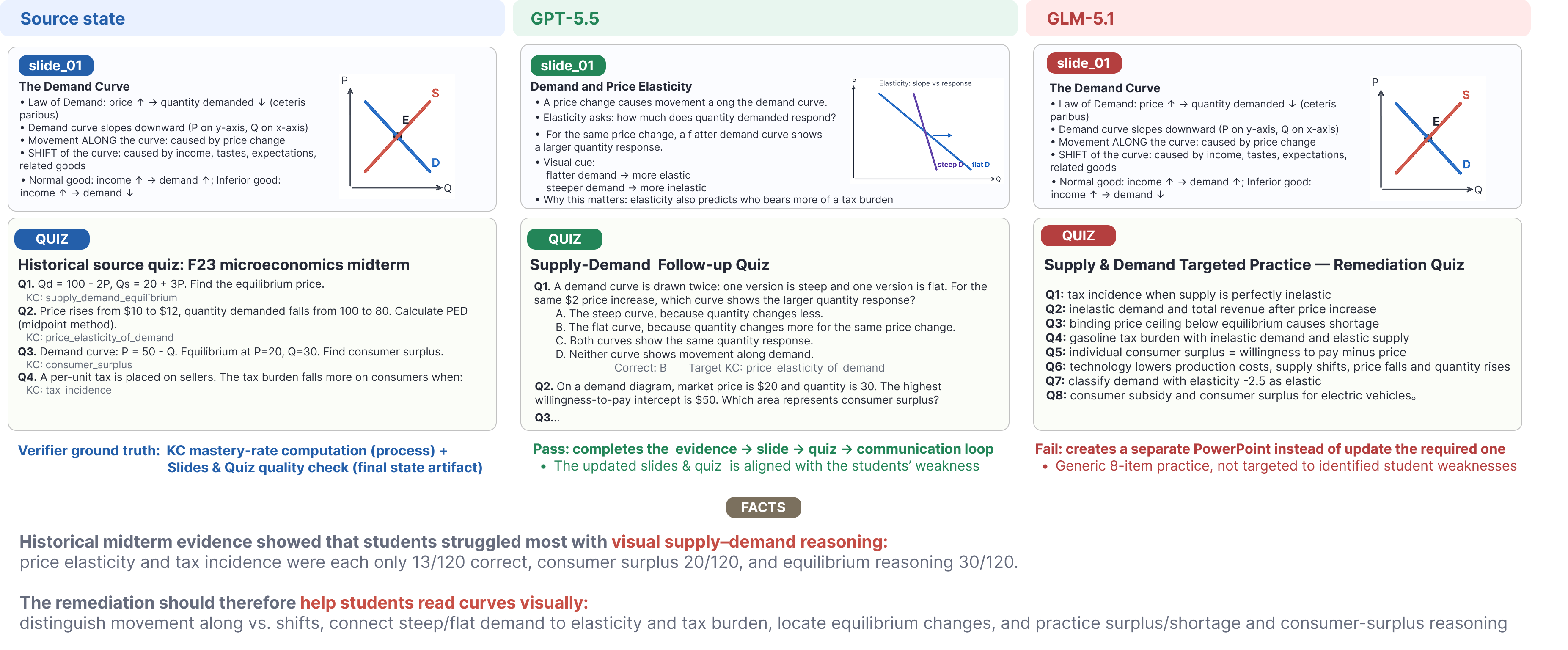}
\caption{\rev{\textbf{Artifact-level contrast in \texttt{MM-04}.} The source state defines both the diagnosed need and the object that must change. GPT-5.5 preserves both constraints in a targeted slide-and-quiz intervention; GLM-5.1 produces plausible materials but breaks object identity and alignment to the diagnosed weakness.}}
\label{fig:slide_case}
\end{figure}

\rev{Together, the cases show why local competence is not automatically compositional. In Stage~2, the warranted mechanism must pass from tutor prompts into the learner's own reasoning without premature disclosure; in Stage~3, the instructor's objective must pass from course evidence through diagnosis and object identity to correctly ordered, persistent action. Matched checks make each break predeclared and observable rather than inferred from the final output: a final-answer judge would miss tutor-supplied interpretation and the untested handoff in \texttt{S0-54}, while a tool-completion judge would miss relevant content written into the wrong artifact in \texttt{MM-04}.}

\section{\rev{Discussion}}
\label{sec:discussion}

Our results expose a central tension: the behavior that makes an agent useful as an assistant can make it ineffective as a tutor. General-purpose assistants are typically optimized to resolve requests quickly, whereas teaching sometimes requires withholding an explanation, resisting a shortcut, allowing productive struggle, and returning intellectual control to the learner. The strategic-surface profile was the lowest-scoring Stage 2 condition for 15 of 17 models, suggesting that over-helpfulness is a practical failure mode for current tutor agents. Future systems may therefore need objectives that reward learner-generated reasoning, calibrated support, and verification of understanding rather than immediate answer delivery or conversational satisfaction.

The transition from tutor to institutional actor creates a different challenge. Once an agent can grade work, revise course materials, or contact students, pedagogical quality becomes inseparable from authority, privacy, and accountability. The low performance on long-chain workflows is consistent with a gap between producing plausible local actions and completing a warranted institutional intervention. In deployment, tutor agents should therefore operate as bounded delegates: consequential actions should retain their supporting evidence, expose what was changed and why, verify the resulting state, and seek approval when evidence or authority is insufficient. Greater autonomy without these properties would expand the consequences of an error rather than the agent’s teaching capability.

Finally, the observed rank reshuffling across stages cautions against treating teaching ability as a single property of a base model. Pedagogical judgment, adaptive interaction, and institutional execution require different mechanisms and safeguards. The practical unit of deployment is therefore the tutor system, combining a model with learner-state tracking, pedagogical policy, action controls, verification, and human oversight. \systemName{} does not determine which architecture is best, but it makes it possible to test whether such systems become more learning-oriented without becoming less operationally reliable.

\Needspace{11\baselineskip}
\section{Limitations}
\label{sec:limitations-release}

\textsc{\systemName{}} provides inspectable, task-level evidence about whether agents preserve a bounded teaching contract. Its controlled design uses simulated learners in Stage~2 and a deterministic synthetic LMS in Stage~3; it spans multiple disciplines but remains weighted toward STEM. These choices enable controlled comparison and localized diagnosis. Future work should connect performance on these controlled episodes to longitudinal learning outcomes, extend coverage to more diverse subjects and learner populations, and evaluate tutor agents under real classroom and institutional constraints.



\begin{ack}
\noindent\textbf{Funding and Competing interests.}
The authors received no specific funding for this work and declare no additional competing interests.

\end{ack}

\bibliographystyle{plainnat}
\bibliography{references}

@article{shulman1986those,
  title={Those who understand: Knowledge growth in teaching},
  author={Shulman, Lee S.},
  journal={Educational Researcher},
  volume={15},
  number={2},
  pages={4--14},
  year={1986}
}

@article{ball2008content,
  title={Content knowledge for teaching: What makes it special?},
  author={Ball, Deborah Loewenberg and Thames, Mark Hoover and Phelps, Geoffrey},
  journal={Journal of Teacher Education},
  volume={59},
  number={5},
  pages={389--407},
  year={2008}
}

@article{black1998assessment,
  title={Assessment and classroom learning},
  author={Black, Paul and Wiliam, Dylan},
  journal={Assessment in Education},
  volume={5},
  number={1},
  pages={7--74},
  year={1998}
}

@book{vygotsky1978mind,
  title={Mind in Society: The Development of Higher Psychological Processes},
  author={Vygotsky, Lev S.},
  year={1978},
  publisher={Harvard University Press}
}

@article{wood1976role,
  title={The role of tutoring in problem solving},
  author={Wood, David and Bruner, Jerome S. and Ross, Gail},
  journal={Journal of Child Psychology and Psychiatry},
  volume={17},
  number={2},
  pages={89--100},
  year={1976}
}

@article{stiggins2002assessment,
  title={Assessment crisis: The absence of assessment FOR learning},
  author={Stiggins, Richard J.},
  journal={Phi Delta Kappan},
  volume={83},
  number={10},
  pages={758--765},
  year={2002}
}

@article{mandinach2016data,
  title={What does it mean for teachers to be data literate?},
  author={Mandinach, Ellen B. and Gummer, Edith S.},
  journal={Educational Researcher},
  volume={45},
  number={6},
  pages={366--376},
  year={2016}
}

@book{branch2009instructional,
  title={Instructional Design: The ADDIE Approach},
  author={Branch, Robert Maribe},
  publisher={Springer},
  year={2009}
}

@article{zimmerman2002self,
  title={Becoming a self-regulated learner: An overview},
  author={Zimmerman, Barry J.},
  journal={Theory Into Practice},
  volume={41},
  number={2},
  pages={64--70},
  year={2002}
}

@inproceedings{Zhou2023BRFE,
  author    = {Qiqi Zhou and Yanyan Shen and Lei Chen},
  title     = {Narrow the Input Mismatch in Deep Graph Neural Network Distillation},
  booktitle = {Proceedings of the 29th ACM SIGKDD Conference on Knowledge Discovery and Data Mining},
  pages     = {3581--3592},
  year      = {2023},
  publisher = {ACM},
  doi       = {10.1145/3580305.3599442}
}

@article{Zhou2025PFNC,
  author  = {Qiqi Zhou and Yanyan Shen and Lei Chen},
  title   = {Faster Convergence in Mini-batch Graph Neural Networks Training with Pseudo Full Neighborhood Compensation},
  journal = {Proceedings of the VLDB Endowment},
  volume  = {18},
  number  = {11},
  pages   = {4309--4322},
  year    = {2025},
  doi     = {10.14778/3749646.3749695}
}

@inproceedings{sheng2026,
author = {Sheng, Rui and Yang, Yukun and Shi, Chuhan and Lin, Yanna and Chen, Zixin and Qu, Huamin and Cheng, Furui},
title = {DiLLS: Interactive Diagnosis of LLM-based Multi-agent Systems via Layered Summary of Agent Behaviors},
year = {2026},
isbn = {9798400722783},
publisher = {Association for Computing Machinery},
url = {https://doi.org/10.1145/3772318.3790815},
doi = {10.1145/3772318.3790815},
booktitle = {Proceedings of the 2026 CHI Conference on Human Factors in Computing Systems},
articleno = {1077},
numpages = {17},
location = {Barcelona, Spain},
series = {CHI '26}
}

@inproceedings{dong2026,
author = {Dong, Haoyu and Sheng, Rui and Zhang, Shuhao and Sun, Yushi and Wu, Dingyang and Chao, Hanxiang and Isayev, Olexandr and Qu, Huamin and Wu, Yuyang and Lin, Yanna},
title = {MolecularCanvas: LLM-assisted Small-Molecule Drug Discovery via Structure-Guided Constraints},
year = {2026},
isbn = {9798400728563},
publisher = {Association for Computing Machinery},
url = {https://doi.org/10.1145/3830398.3830585},
doi = {10.1145/3830398.3830585},
booktitle = {Proceedings of the 39th Annual ACM Symposium on User Interface Software and Technology},
articleno = {238},
numpages = {13},
series = {UIST '26}
}

@article{sun2026,
author = {Sun, Yushi and Xin, Hao and Sun, Kai and Xu, Yifan Ethan and Yang, Xiao and Dong, Xin Luna and Tang, Nan and Chen, Lei},
title = {Are Large Language Models a Good Replacement of Taxonomies?},
year = {2024},
issue_date = {July 2024},
publisher = {VLDB Endowment},
volume = {17},
number = {11},
issn = {2150-8097},
url = {https://doi.org/10.14778/3681954.3681973},
doi = {10.14778/3681954.3681973},
journal = {Proc. VLDB Endow.},
month = jul,
pages = {2919–2932},
numpages = {14}
}

@article{yang2024crag,
  title={Crag-comprehensive rag benchmark},
  author={Yang, Xiao and Sun, Kai and Xin, Hao and Sun, Yushi and Bhalla, Nikita and Chen, Xiangsen and Choudhary, Sajal and Gui, Rongze D and Jiang, Ziran W and Jiang, Ziyu and others},
  journal={Advances in Neural Information Processing Systems},
  volume={37},
  pages={10470--10490},
  year={2024}
}

@inproceedings{macina2023mathdial,
  title={MathDial: A Dialogue Tutoring Dataset with Rich Pedagogical Properties Grounded in Math Reasoning Problems},
  author={Macina, Jakub and Daheim, Nico and Pal Chowdhury, Sankalan and Sinha, Tanmay and Kapur, Manu and Gurevych, Iryna and Sachan, Mrinmaya},
  booktitle={Findings of the Association for Computational Linguistics: EMNLP 2023},
  pages={5602--5621},
  year={2023},
  doi={10.18653/v1/2023.findings-emnlp.372}
}

@article{vosniadou1992mental,
  title={Mental Models of the Earth: A Study of Conceptual Change in Childhood},
  author={Vosniadou, Stella and Brewer, William F.},
  journal={Cognitive Psychology},
  volume={24},
  number={4},
  pages={535--585},
  year={1992},
  doi={10.1016/0010-0285(92)90018-W}
}

@article{hendrycks2021math,
  title={Measuring mathematical problem solving with the MATH dataset},
  author={Hendrycks, Dan and Burns, Collin and others},
  journal={NeurIPS},
  year={2021}
}

@article{haladyna1993,
  title={How many options is enough for a multiple-choice test item?},
  author={Haladyna, Thomas M. and Downing, Steven M.},
  journal={Educational and Psychological Measurement},
  volume={53},
  number={4},
  pages={999--1010},
  year={1993}
}

@article{krathwohl2002revision,
  title={A Revision of Bloom's Taxonomy: An Overview},
  author={Krathwohl, David R.},
  journal={Theory Into Practice},
  volume={41},
  number={4},
  pages={212--218},
  year={2002},
  doi={10.1207/s15430421tip4104_2}
}

@article{gigerenzer1995,
  title={How to improve Bayesian reasoning without instruction: Frequency formats},
  author={Gigerenzer, Gerd and Hoffrage, Ulrich},
  journal={Psychological Review},
  volume={102},
  number={4},
  pages={684--704},
  year={1995}
}

@article{wood1975assisted,
  title={A study of assisted problem-solving},
  author={Wood, David and Middleton, David},
  journal={British Journal of Psychology},
  volume={66},
  number={2},
  pages={181--191},
  year={1975}
}

@article{vandepol2010scaffolding,
  title={Scaffolding in Teacher--Student Interaction: A Decade of Research},
  author={van de Pol, Janneke and Volman, Monique and Beishuizen, Jos},
  journal={Educational Psychology Review},
  volume={22},
  number={3},
  pages={271--296},
  year={2010}
}

@article{brown1978diagnostic,
  title={Diagnostic Models for Procedural Bugs in Basic Mathematical Skills},
  author={Brown, John Seely and Burton, Richard R.},
  journal={Cognitive Science},
  volume={2},
  number={2},
  pages={155--192},
  year={1978}
}

@article{sfard1991dual,
  title={On the Dual Nature of Mathematical Conceptions: Reflections on Processes and Objects as Different Sides of the Same Coin},
  author={Sfard, Anna},
  journal={Educational Studies in Mathematics},
  volume={22},
  number={1},
  pages={1--36},
  year={1991}
}

@article{simpson1951interpretation,
  title={The Interpretation of Interaction in Contingency Tables},
  author={Simpson, Edward H.},
  journal={Journal of the Royal Statistical Society: Series B},
  volume={13},
  number={2},
  pages={238--241},
  year={1951}
}

@article{corbett1994knowledge,
  title={Knowledge Tracing: Modeling the Acquisition of Procedural Knowledge},
  author={Corbett, Albert T. and Anderson, John R.},
  journal={User Modeling and User-Adapted Interaction},
  volume={4},
  number={4},
  pages={253--278},
  year={1994},
  doi={10.1007/BF01099821}
}

@article{sweller1988cognitive,
  title={Cognitive Load During Problem Solving: Effects on Learning},
  author={Sweller, John},
  journal={Cognitive Science},
  volume={12},
  number={2},
  pages={257--285},
  year={1988},
  doi={10.1207/s15516709cog1202_4}
}

@article{chi2014icap,
  title={The {ICAP} Framework: Linking Cognitive Engagement to Active Learning Outcomes},
  author={Chi, Michelene T. H. and Wylie, Ruth},
  journal={Educational Psychologist},
  volume={49},
  number={4},
  pages={219--243},
  year={2014},
  doi={10.1080/00461520.2014.965823}
}

@article{ryan2000intrinsic,
  title={Intrinsic and Extrinsic Motivations: Classic Definitions and New Directions},
  author={Ryan, Richard M. and Deci, Edward L.},
  journal={Contemporary Educational Psychology},
  volume={25},
  number={1},
  pages={54--67},
  year={2000},
  doi={10.1006/ceps.1999.1020}
}

@article{pekrun2006control,
  title={The Control-Value Theory of Achievement Emotions: Assumptions, Corollaries, and Implications for Educational Research and Practice},
  author={Pekrun, Reinhard},
  journal={Educational Psychology Review},
  volume={18},
  number={4},
  pages={315--341},
  year={2006},
  doi={10.1007/s10648-006-9029-9}
}

@article{goldberg1990alternative,
  title={An Alternative ``Description of Personality'': The Big-Five Factor Structure},
  author={Goldberg, Lewis R.},
  journal={Journal of Personality and Social Psychology},
  volume={59},
  number={6},
  pages={1216--1229},
  year={1990},
  doi={10.1037/0022-3514.59.6.1216}
}

@article{duckworth2007grit,
  title={Grit: Perseverance and Passion for Long-Term Goals},
  author={Duckworth, Angela L. and Peterson, Christopher and Matthews, Michael D. and Kelly, Dennis R.},
  journal={Journal of Personality and Social Psychology},
  volume={92},
  number={6},
  pages={1087--1101},
  year={2007},
  doi={10.1037/0022-3514.92.6.1087}
}

@inproceedings{wu2025imperfection,
  title={Embracing Imperfection: Simulating Students with Diverse Cognitive Levels Using {LLM}-based Agents},
  author={Wu, Tao and Chen, Jingyuan and Lin, Wang and Li, Mengze and Zhu, Yumeng and Li, Ang and Kuang, Kun and Wu, Fei},
  booktitle={Proceedings of the 63rd Annual Meeting of the Association for Computational Linguistics (Volume 1: Long Papers)},
  pages={9887--9908},
  year={2025},
  doi={10.18653/v1/2025.acl-long.488}
}

@article{biggs1996enhancing,
  title={Enhancing teaching through constructive alignment},
  author={Biggs, John},
  journal={Higher Education},
  volume={32},
  pages={347--364},
  year={1996}
}

@article{wang2024mmlupro,
  title={{MMLU-Pro}: A More Robust and Challenging Multi-Task Language Understanding Benchmark},
  author={Wang, Yubo and Ma, Xueguang and Zhang, Ge and Ni, Yuansheng and Chandra, Abhranil and Guo, Shiguang and Ren, Weiming and Arulraj, Aaran and He, Xuan and Jiang, Ziyan and Li, Tianle and Ku, Max and Wang, Kai and Zhuang, Alex and Fan, Rongqi and Yue, Xiang and Chen, Wenhu},
  journal={arXiv preprint arXiv:2406.01574},
  year={2024}
}

@misc{mitocw,
  title={{MIT OpenCourseWare}},
  author={{Massachusetts Institute of Technology}},
  howpublished={\url{https://ocw.mit.edu/}},
  year={2026},
  note={Accessed May 1, 2026}
}

@misc{ho2025verilogcoder,
  title={{VerilogCoder}: Autonomous Verilog Coding Agents with Graph-Based Planning and Abstract Syntax Tree-Based Waveform Tracing Tool},
  author={Ho, Chia-Tung and Ren, Haoxing and Khailany, Brucek},
  year={2024},
  eprint={2408.08927},
  archivePrefix={arXiv},
  primaryClass={cs.AI}
}

@misc{islam2025codesim,
  title={{CODESIM}: Multi-Agent Code Generation and Problem Solving through Simulation-Driven Planning and Debugging},
  author={Islam, Md. Ashraful and Ali, Mohammed Eunus and Parvez, Md Rizwan},
  year={2025},
  eprint={2502.05664},
  archivePrefix={arXiv},
  primaryClass={cs.CL}
}

@misc{wang2025medkgi,
  title={{MedKGI}: Iterative Differential Diagnosis with Medical Knowledge Graphs and Information-Guided Inquiring},
  author={Wang, Qipeng and Sheng, Rui and Li, Yafei and Qu, Huamin and Sun, Yushi and Zhu, Min},
  year={2025},
  eprint={2512.24181},
  archivePrefix={arXiv},
  primaryClass={cs.AI}
}

@misc{ghafarollahi2025sciagents,
  title={{SciAgents}: Automating Scientific Discovery through Multi-Agent Intelligent Graph Reasoning},
  author={Ghafarollahi, Alireza and Buehler, Markus J.},
  year={2024},
  eprint={2409.05556},
  archivePrefix={arXiv},
  primaryClass={cs.AI}
}

@inproceedings{hendrycks2020mmlu,
  title        = {Measuring Massive Multitask Language Understanding},
  author       = {Hendrycks, Dan and Burns, Collin and Basart, Steven and Zou, Andy and Mazeika, Mantas and Song, Dawn and Steinhardt, Jacob},
  booktitle    = {International Conference on Learning Representations},
  year         = {2021},
  url          = {https://openreview.net/forum?id=d7KBjmI3GmQ}
}

@inproceedings{hou2024eeval,
  title        = {{E}-{EVAL}: A Comprehensive {C}hinese K-12 Education Evaluation Benchmark for Large Language Models},
  author       = {Hou, Jinchang and Ao, Chang and Wu, Haihong and Kong, Xiangtao and Zheng, Zhigang and Tang, Daijia and Li, Chengming and Hu, Xiping and Xu, Ruifeng and Ni, Shiwen and Yang, Min},
  booktitle    = {Findings of the Association for Computational Linguistics: ACL 2024},
  year         = {2024},
  month        = aug,
  address      = {Bangkok, Thailand},
  publisher    = {Association for Computational Linguistics},
  pages        = {7753--7774},
  doi          = {10.18653/v1/2024.findings-acl.462},
  url          = {https://aclanthology.org/2024.findings-acl.462/}
}

@misc{ma2025edueval,
  title         = {{EduEval}: A Hierarchical Cognitive Benchmark for Evaluating Large Language Models in Chinese Education},
  author        = {Ma, Guoqing and Zhu, Jia and Guo, Hanghui and Shi, Weijie and Cui, Yue and Shen, Jiawei and Li, Zilong and Liang, Yidan},
  year          = {2025},
  eprint        = {2512.00290},
  archivePrefix = {arXiv},
  primaryClass  = {cs.CL},
  url           = {https://arxiv.org/abs/2512.00290}
}

@misc{xu2025edubench,
  title         = {{EduBench}: A Comprehensive Benchmarking Dataset for Evaluating Large Language Models in Diverse Educational Scenarios},
  author        = {Xu, Bin and Bai, Yu and Sun, Huashan and Lin, Yiguan and Liu, Siming and Liang, Xinyue and Li, Yaolin and Gao, Yang and Huang, Heyan},
  year          = {2025},
  eprint        = {2505.16160},
  archivePrefix = {arXiv},
  primaryClass  = {cs.CL},
  url           = {https://arxiv.org/abs/2505.16160}
}

@inproceedings{macina2025mathtutorbench,
  title        = {{M}ath{T}utor{B}ench: A Benchmark for Measuring Open-ended Pedagogical Capabilities of {LLM} Tutors},
  author       = {Macina, Jakub and Daheim, Nico and Hakimi, Ido and Kapur, Manu and Gurevych, Iryna and Sachan, Mrinmaya},
  booktitle    = {Proceedings of the 2025 Conference on Empirical Methods in Natural Language Processing},
  year         = {2025},
  month        = nov,
  address      = {Suzhou, China},
  publisher    = {Association for Computational Linguistics},
  pages        = {204--221},
  doi          = {10.18653/v1/2025.emnlp-main.11},
  url          = {https://aclanthology.org/2025.emnlp-main.11/}
}

@misc{srinivasa2025tutorbench,
  title         = {{TutorBench}: A Benchmark To Assess Tutoring Capabilities Of Large Language Models},
  author        = {Srinivasa, Rakshith Sharma and Che, Zora and Zhang, Chen Bo Calvin and Buendia, Diego A. Mares and Montoya, Ernesto Gabriel Hern{\'a}ndez and Park, Jayeon and Lee, Dean and Mangialardi, Guillermo A. and Ng, Charmaine and Hernandez-Cardona, Ed-Yeremai and Gunjal, Anisha and He, Yunzhong and Liu, Bing and Xing, Chen},
  year          = {2025},
  eprint        = {2510.02663},
  archivePrefix = {arXiv},
  primaryClass  = {cs.CL},
  url           = {https://arxiv.org/abs/2510.02663}
}

@misc{learnlm2024,
  title         = {{LearnLM}: Improving {Gemini} for Learning},
  author        = {{LearnLM Team}},
  year          = {2024},
  eprint        = {2412.16429},
  archivePrefix = {arXiv},
  primaryClass  = {cs.CL},
  url           = {https://arxiv.org/abs/2412.16429}
}

@article{feng2009assistments,
  title        = {Addressing the Assessment Challenge with an Online System That Tutors as It Assesses},
  author       = {Feng, Mingyu and Heffernan, Neil T. and Koedinger, Kenneth R.},
  journal      = {User Modeling and User-Adapted Interaction},
  volume       = {19},
  number       = {3},
  pages        = {243--266},
  year         = {2009},
  publisher    = {Springer},
  doi          = {10.1007/s11257-009-9063-7}
}

@misc{choi2020ednet,
  title         = {{EdNet}: A Large-Scale Hierarchical Dataset in Education},
  author        = {Choi, Youngduck and Lee, Youngnam and Shin, Dongmin and Cho, Junghyun and Park, Seoyon and Lee, Seewoo and Baek, Jineon and Bae, Chan and Kim, Byungsoo and Heo, Jaewe},
  year          = {2020},
  eprint        = {1912.03072},
  archivePrefix = {arXiv},
  primaryClass  = {cs.CY},
  url           = {https://arxiv.org/abs/1912.03072}
}

@inproceedings{yu2021mooccubex,
  title        = {{MOOCCubeX}: A Large Knowledge-centered Repository for Adaptive Learning in {MOOC}s},
  author       = {Yu, Jifan and Wang, Yuquan and Zhong, Qingyang and Luo, Gan and Mao, Yiming and Sun, Kai and Feng, Wenzheng and Xu, Wei and Cao, Shulin and Zeng, Kaisheng and Yao, Zijun and Hou, Lei and Lin, Yankai and Li, Peng and Zhou, Jie and Xu, Bin and Li, Juanzi and Tang, Jie and Sun, Maosong},
  booktitle    = {Proceedings of the 30th ACM International Conference on Information \& Knowledge Management},
  year         = {2021},
  pages        = {4643--4652},
  publisher    = {Association for Computing Machinery},
  doi          = {10.1145/3459637.3482010},
  url          = {https://doi.org/10.1145/3459637.3482010}
}

@inproceedings{yao2024taubench,
  title        = {{$\tau$}-bench: A Benchmark for Tool-Agent-User Interaction in Real-World Domains},
  author       = {Yao, Shunyu and Shinn, Noah and Razavi, Pedram and Narasimhan, Karthik},
  booktitle    = {International Conference on Learning Representations},
  year         = {2025},
  url          = {https://openreview.net/forum?id=roNSXZpUDN}
}

@misc{xu2024theagentcompany,
  title         = {{TheAgentCompany}: Benchmarking {LLM} Agents on Consequential Real World Tasks},
  author        = {Xu, Frank F. and Song, Yufan and Li, Boxuan and Tang, Yuxuan and Jain, Kritanjali and Bao, Mengxue and Wang, Zora Z. and Zhou, Xuhui and Guo, Zhitong and Cao, Murong and Yang, Mingyang and Lu, Hao Yang and Martin, Amaad and Su, Zhe and Maben, Leander Melroy and Mehta, Raj and Chi, Wayne and Jang, Lawrence and Xie, Yiqing and Zhou, Shuyan and Neubig, Graham},
  year          = {2024},
  eprint        = {2412.14161},
  archivePrefix = {arXiv},
  primaryClass  = {cs.AI},
  url           = {https://arxiv.org/abs/2412.14161}
}

@article{chen2026informal,
  title={Informal Learning Emerges in Everyday Human-LLM Interaction},
  author={Chen, Zixin and Li, Haotian and Xiao, Ziang and Qu, Huamin and Xie, Xing},
  journal={arXiv preprint arXiv:2607.17643},
  year={2026}
}

@inproceedings{chen2026passive,
  title={From Passive Consumption to Active Interaction: Exploring Interactive LLM Scaffolding to Support Learning Engagement},
  author={Chen, Zixin and Li, Haotian and Liu, Zhe and Qu, Huamin and Xie, Xing},
  booktitle={Proceedings of the Extended Abstracts of the 2026 CHI Conference on Human Factors in Computing Systems},
  pages={1--6},
  year={2026}
}

@misc{li2025toolathlon,
  title         = {The Tool Decathlon: Benchmarking Language Agents for Diverse, Realistic, and Long-Horizon Task Execution},
  author        = {Li, Junlong and Zhao, Wenshuo and Zhao, Jian and Zeng, Weihao and Wu, Haoze and Wang, Xiaochen and Ge, Rui and Cao, Yuxuan and Huang, Yuzhen and Liu, Wei and Liu, Junteng and Su, Zhaochen and Guo, Yiyang and Zhou, Fan and Zhang, Lueyang and Michelini, Juan and Wang, Xingyao and Yue, Xiang and Zhou, Shuyan and Neubig, Graham and He, Junxian},
  year          = {2025},
  eprint        = {2510.25726},
  archivePrefix = {arXiv},
  primaryClass  = {cs.CL},
  url           = {https://arxiv.org/abs/2510.25726}
}

@book{danielson2007enhancing,
  title={Enhancing professional practice: A framework for teaching},
  author={Danielson, Charlotte},
  year={2007},
  publisher={AsCD}
}

@article{dmello2012dynamics,
  title={Dynamics of affective states during complex learning},
  author={D'Mello, Sidney and Graesser, Arthur},
  journal={Learning and Instruction},
  volume={22},
  number={2},
  pages={145--157},
  year={2012}
}

@article{chen2026vizqstudio,
  title={VizQStudio: Iterative Visualization Literacy MCQs Design with Simulated Students},
  author={Chen, Zixin and Zeng, Yuhang and Song, Sicheng and Lin, Yanna and Xu, Xian and Qu, Huamin and Xia, Meng},
  journal={IEEE Transactions on Visualization and Computer Graphics},
  year={2026},
  publisher={IEEE}
}

@inproceedings{chen2025cograder,
  title={CoGrader: Transforming Instructors' Assessment of Project Reports through Collaborative LLM Integration},
  author={Chen, Zixin and Wang, Jiachen and Li, Yumeng and Li, Haobo and Shi, Chuhan and Zhang, Rong and Qu, Huamin},
  booktitle={Proceedings of the 38th Annual ACM Symposium on User Interface Software and Technology},
  pages={1--18},
  year={2025}
}

@article{chen2024stugptviz,
  title={StuGPTViz: A visual analytics approach to understand student-ChatGPT interactions},
  author={Chen, Zixin and Wang, Jiachen and Xia, Meng and Shigyo, Kento and Liu, Dingdong and Zhang, Rong and Qu, Huamin},
  journal={IEEE Transactions on Visualization and Computer Graphics},
  volume={31},
  number={1},
  pages={908--918},
  year={2024},
  publisher={IEEE}
}

\clearpage
\appendix
\Needspace{5\baselineskip}
\section{Related Benchmarks and the Missing Teaching-System Evaluation}
\label{sec:related-positioning}

\paragraph{\rev{Educational competence, tutoring, and learner modeling.}}
\rev{Educational QA and exam-style benchmarks such as MATH, MMLU, E-EVAL, EduEval, and EduBench evaluate subject-matter competence and educational problem solving~\citep{hendrycks2021math,hendrycks2020mmlu,hou2024eeval,ma2025edueval,xu2025edubench}. Tutoring resources such as MathDial, MathTutorBench, TutorBench, and LearnLM evaluations move closer to instruction by assessing hints, feedback, explanations, and dialogue behavior~\citep{macina2023mathdial,macina2025mathtutorbench,srinivasa2025tutorbench,learnlm2024}. Student-modeling and learning-analytics resources such as ASSISTments, EdNet, and MOOCCubeX instead estimate learner knowledge, engagement, or performance from educational records~\citep{feng2009assistments,choi2020ednet,yu2021mooccubex}. Together, these lines supply content, response-quality, and learner-state signals. Complementary research on pedagogical content knowledge, formative assessment, assessment literacy, and teacher data literacy characterizes how teachers turn such evidence into instructional decisions~\citep{shulman1986those,ball2008content,black1998assessment,stiggins2002assessment,mandinach2016data}. Estimating state or scoring a local response therefore does not test the full transition from learner evidence to adaptive teaching. Research on graph neural networks has addressed receptive-field mismatch in distillation~\citep{Zhou2023BRFE} and sampling-induced gradient error in mini-batch training~\citep{Zhou2025PFNC}. In \textsc{\systemName{}}, prerequisite graphs serve as evidence for a downstream instructional decision: identifying which concept to remediate.}

\paragraph{\rev{Agentic workflow evaluation.}}
\rev{General agent benchmarks such as $\tau$-bench, TheAgentCompany, and Toolathlon evaluate planning, tool use, persistent state, and long-horizon execution~\citep{yao2024taubench,xu2024theagentcompany,li2025toolathlon}. These capabilities are necessary for teacher-facing agents, but their success criteria do not ordinarily ask whether the resulting action is pedagogically warranted by its source evidence. Teaching-workflow evaluation must therefore verify continuity across evidence, decision, action, and resulting state.}

\begin{table}[H]
\centering
\scriptsize
\setlength{\tabcolsep}{2.3pt}
\renewcommand{\arraystretch}{1.00}

\resizebox{\textwidth}{!}{%
\begin{tabular}{@{}llcccccccc@{}}
\toprule
\textbf{Benchmark / resource} &
\textbf{Unit} &
\shortstack{\textbf{Subject}\\\textbf{content}} &
\shortstack{\textbf{Local tutor}\\\textbf{response}} &
\shortstack{\textbf{Full tutoring}\\\textbf{trajectory}} &
\shortstack{\textbf{Teacher}\\\textbf{judgment / PCK}} &
\shortstack{\textbf{LMS / tool}\\\textbf{action}} &
\shortstack{\textbf{Pedagogy-}\\\textbf{constrained}\\\textbf{action}} &
\shortstack{\textbf{State / process}\\\textbf{checks}} &
\shortstack{\textbf{Learning}\\\textbf{outcome}} \\
\midrule

\multicolumn{10}{@{}c@{}}{\emph{Educational QA and exam-style benchmarks}} \\
\midrule
MATH      & Problem      & \eabyes & \eabno   & \eabno   & \eabno   & \eabno  & \eabno  & \eabno   & \eabno \\
MMLU      & Exam item    & \eabyes & \eabno   & \eabno   & \eabno   & \eabno  & \eabno  & \eabno   & \eabno \\
E-EVAL    & Exam item    & \eabyes & \eabpart & \eabno   & \eabpart & \eabno  & \eabno  & \eabno   & \eabno \\
EduEval   & Edu scenario & \eabyes & \eabpart & \eabno   & \eabpart & \eabno  & \eabno  & \eabno   & \eabno \\
EduBench  & Edu scenario & \eabyes & \eabpart & \eabno   & \eabpart & \eabno  & \eabno  & \eabno   & \eabno \\
\midrule

\multicolumn{10}{@{}c@{}}{\emph{Tutoring-dialogue and local tutor-response benchmarks}} \\
\midrule
MathDial       & Dialogue       & \eabpart & \eabyes & \eabpart & \eabpart & \eabno & \eabpart & \eabno   & \eabno \\
MathTutorBench & Tutor resp.    & \eabpart & \eabyes & \eabpart & \eabpart & \eabno & \eabpart & \eabpart & \eabpart \\
TutorBench     & Tutor resp.    & \eabpart & \eabyes & \eabpart & \eabpart & \eabno & \eabpart & \eabpart & \eabpart \\
LearnLM evals  & Tutor scenario & \eabpart & \eabyes & \eabpart & \eabpart & \eabno & \eabpart & \eabpart & \eabpart \\
\midrule

\multicolumn{10}{@{}c@{}}{\emph{Student modeling and learning-analytics resources}} \\
\midrule
ASSISTments & Learner log    & \eabpart & \eabno & \eabno & \eabpart & \eabno & \eabno & \eabpart & \eabyes \\
EdNet       & Learner log    & \eabpart & \eabno & \eabno & \eabpart & \eabno & \eabno & \eabpart & \eabyes \\
MOOCCubeX   & Learning graph & \eabpart & \eabno & \eabno & \eabpart & \eabno & \eabno & \eabpart & \eabpart \\
\midrule

\multicolumn{10}{@{}c@{}}{\emph{General agent and tool-use benchmarks}} \\
\midrule
$\tau$-bench    & Tool task      & \eabno & \eabno & \eabno & \eabno & \eabyes & \eabno & \eabyes & \eabno \\
TheAgentCompany & Workplace task & \eabno & \eabno & \eabno & \eabno & \eabyes & \eabno & \eabyes & \eabno \\
Toolathlon      & Tool workflow  & \eabno & \eabno & \eabno & \eabno & \eabyes & \eabno & \eabyes & \eabno \\
\midrule

\multicolumn{10}{@{}c@{}}{\emph{Teaching-system benchmark}} \\
\midrule
\textsc{\systemName{}} & Teaching episode
& \eabyes & \eabyes & \eabyes & \eabyes & \eabyes & \eabyes & \eabyes & \eabyes \\
\bottomrule
\end{tabular}%
}
\caption{\textbf{Positioning relative to representative educational and agent benchmarks.}
\eabyes{} denotes a primary target, \eabpart{} partial or indirect coverage, and \eabno{} no primary coverage. Prior resources cover important ingredients, but \textsc{\systemName{}} targets the missing conjunction required for \rev{complete teaching-system evaluation}: pedagogical judgment, situated tutoring, and institutional teaching action under educational constraints. }
\label{tab:related-benchmarks}
\end{table}

\paragraph{\rev{The missing teaching-system episode.}}
\rev{Prior work evaluates individual components of educational agency, not their conjunction. We therefore define an auditable \emph{teaching-system episode}: evidence supports a teacher decision, the decision guides productive interaction or execution, and the intended learner or course consequence is observable. \textsc{\systemName{}} decomposes this object into evidence-grounded judgment (Stage~1), situated tutoring and learner-produced evidence (Stage~2), and end-to-end LMS workflows (Stage~3), preserving matched evidence that identifies breaks in the teaching chain.}

\section{\rev{Stage 1 Design Audit and Representative Cases}}
\label{app:stage1-design-audit}
\begingroup

Stage~1 contains 117 pedagogical-judgment tasks. Table~\ref{tab:stage1-source-strata} reports the provenance of their evidence and ground truth; Table~\ref{tab:stage1-decision-spans} groups them by the object of the required decision. For mixed prompts, the task-specific verifier determines the assignment: learner-state inference when the scored commitment is an explanatory learner/KC mechanism, instructional-action selection when it is a particular teaching move, and assessment and claim audit when it is a verdict on an assessment artifact or empirical conclusion.

For auditability, the third family retains two subtypes in the task map: assessment-artifact audit (8 tasks) and empirical-claim audit (8). Professional-knowledge tags may cross the three families: a teaching move can require misconception diagnosis, but it is scored as action selection; an audit can propose a repair, but it is scored first on whether the artifact or claim is defensible. Task-ID assignments are provided in \texttt{audits/stage1\_task\_map.tsv}.

\begin{table*}[h]
\caption{\rev{\textbf{Stage~1 source strata.} Stage~1 asks the model to infer teacher-like conclusions from evidence without dialogue management or tool execution. Counts are audited against the current 117-task set.}}
\centering
\small
\setlength{\tabcolsep}{4pt}
\renewcommand{\arraystretch}{1.10}
\begin{tabularx}{\textwidth}{>{\raggedright\arraybackslash}p{2.45cm}>{\raggedright\arraybackslash}p{2.9cm}c>{\raggedright\arraybackslash}X}
\toprule
\textbf{Source stratum} & \textbf{Construction basis} & \textbf{Count} & \textbf{Examples of target insights} \\
\midrule
\makecell[l]{MathDial-\\derived} & Source problem, learner work, and teacher-described confusion & 26 & Localize the first conceptual divergence and choose feedback that targets it. \\
\makecell[l]{OER/literature-\\grounded} & Redistributable content plus documented misconceptions or principles & 22 & Diagnose conceptual models and choose representation- or feedback-level interventions. \\
\makecell[l]{MathTutorBench-\\derived} & Fixed tutoring transcripts with known reference reasoning & 17 & Judge whether a tutor move reveals, scaffolds, or verifies the target idea. \\
\makecell[l]{Authored evidence\\contracts} & Deterministic course data or literature-grounded contrasts & 52 & Infer gateway KCs, distinguish surface from conceptual change, and audit assessment validity. \\
\midrule
\textbf{Total} & \textbf{Current Stage~1} & \textbf{117} & Public release includes license-compatible transformed tasks, provenance metadata, and evaluation criteria. \\
\bottomrule
\end{tabularx}
\label{tab:stage1-source-strata}
\end{table*}

\subsection{Source-grounded construction}

MathDial transformations preserve the source problem, learner work, and teacher-described confusion; MathTutorBench transformations preserve the tutoring transcript and reference reasoning about the tutor's move. OER/literature-grounded tasks pair redistributable content with a documented misconception or pedagogical principle. Authored contracts instantiate published mechanisms or deterministic contrasts in item statistics, prerequisite graphs, learner traces, or intervention data. Across all four strata, the target mechanism is fixed by the source evidence, deterministic data, or published theory before authors construct the decision setting, plausible foil, prompt, and matched verifier.

\subsection{Discriminating design and verification}
\label{app:stage1-discriminating}

Stage~1 targets contrastive professional inference rather than content obscurity. Cases pair the warranted decision with a locally plausible shortcut: a visible error instead of its generative misconception, the most complete explanation instead of the learner-appropriate move, low item difficulty as invalidity, an aggregate gap as causal or fair, or score and time-on-task as engagement. A response must commit to the decision and justify its evidence--decision link; a correct calculation, theory name, or generic best practice is insufficient.

The main-text \texttt{EI-AFFECT-01} probe instantiates the affect-dynamics account that unresolved difficulty can deteriorate, whereas difficulty remains productive when the learner uses help, self-corrects, and progresses~\citep{dmello2012dynamics,pekrun2006control}. It holds the content, session, and current score fixed while varying observable process and prior-performance evidence. For the displayed pair, Leo's time rises to 71 seconds per item, with five failed retries and no revision after help; Tom works steadily, uses help independently, self-corrects three times, and improves from 0/10 to 3/10 on the same above-level items. The prompt withholds state labels, so the action must be inferred from the trace.

The complete \texttt{EI-AFFECT-01} contract covers four evidence--action pairs: targeted scaffolding for an active impasse, guided resolution for a deteriorating failure loop, increased challenge or relevance for rapid disengagement, and maintained stretch for productive progress. Five semantic assertions verify these pairings and the resulting non-uniform policy, so a label-only answer or generic ``more practice'' prescription fails.

\subsection{Scoring realization and evaluator audit}
\label{app:stage1-scoring}

All 117 tasks use mechanism-sensitive semantic checks aggregated across multiple evaluator views; 100 also admit a deterministic response anchor. For those 100 tasks,
\begin{equation}
R_1=\begin{cases}
0, & R_{\mathrm{anchor}}=0\ \text{or}\ R_{\mathrm{sem}}=0,\\
0.55R_{\mathrm{anchor}}+0.45R_{\mathrm{sem}}, & \text{otherwise}.
\end{cases}
\end{equation}
The other 17 use semantic assertions alone. Thus a keyword-level conclusion cannot rescue a mechanism-free explanation, while a fluent rationale cannot rescue a missing or contradictory decision.

Evaluator robustness audits test cross-evaluator sensitivity, self-preference, and response-length sensitivity. Together with multi-model aggregation and atomic, mechanism-specific assertions, these checks reduce reliance on one holistic verdict. We therefore emphasize stage profiles and bootstrap-supported contrasts, treating adjacent point estimates as near ties when their difference is small.

\subsection{How the representative cards were selected}

We select cards from the final executable tasks after contract-level review; preliminary review artifacts are used to find defects, not as ground truth. A representative case must satisfy four criteria: (i) the teacher decision is recoverable from cited theory, deterministic data, or source-labeled learner evidence; (ii) a plausible shortcut is present and demonstrably yields a different conclusion; (iii) the ground truth is unique or the acceptable alternatives are stated explicitly; and (iv) response checks and semantic assertions verify the same evidence-to-decision chain. The cards therefore expose task design rather than merely showcasing difficult prompts.

\newcommand{\eabdesignfield}[2]{%
  \par
  \noindent\colorbox{EABStageOne!10}{\strut\textcolor{EABStageOne}{\textbf{#1}}}\hspace{0.45em}#2\par
}

\begingroup
\small
\setlength{\parskip}{0pt}

\subsection{Learner-state inference: \texttt{EI-DIAG-MISC-01}}

\eabdesignfield{Evidence contract}{The learner answers eight multi-digit subtraction items. The authored answer pattern is grounded in the buggy-procedure account of systematic mathematical errors~\citep{brown1978diagnostic}: the rule ``take the larger digit minus the smaller digit in each column; never regroup'' reproduces all eight responses exactly. The two no-regrouping items are correct, so they are positive evidence for the same rule rather than noise.}

\eabdesignfield{Discriminating design}{A generic ``carelessness'' diagnosis cannot explain why correctness coincides exactly with the absence of regrouping. A second tempting diagnosis---difficulty only when subtracting through zero---is refuted by three wrong items containing no zero. The full set, rather than any single error, identifies one mechanism and motivates targeted regrouping instruction instead of more undifferentiated practice.}

\eabdesignfield{Matched verifier}{Checks require the column-wise larger-minus-smaller rule, evidence from the correct no-regrouping items, rejection of the zero-only explanation, and a remediation that confronts the buggy step. Naming a broad ``borrowing problem'' without specifying the executable rule is insufficient.}

\subsection{Assessment and claim audit---assessment subtype: \texttt{EI-C04}}

\eabdesignfield{Evidence contract}{A multiple-choice item asks for the bug in a recursive factorial function. The keyed answer says the base case omits \(n=0\), a second option restates that same defect through \(0! = 1\), and a third correctly notes that negative inputs are unhandled. Responses are nearly uniform and the point-biserial is only \(r_{\mathrm{pb}}=0.11\) (i.e., the item scarcely separates higher- from lower-performing students). The item-writing literature makes the relevant obligation explicit: alternatives must support one intended interpretation of the construct~\citep{haladyna1993}.}

\eabdesignfield{Discriminating design}{The near-uniform split can be misread as four well-functioning options. Instead, three answers are independently defensible unless the stem fixes an input domain, while the remaining distractor is implausible. The statistic therefore reflects an underdetermined construct and overlapping keys, not useful discrimination; the defensible action is to revise the stem rather than celebrate the spread.}

\eabdesignfield{Matched verifier}{Checks require rejecting the ``excellent discrimination'' reading, explaining why the first three options can all be defended, identifying the last option as implausible, and proposing a stem revision that makes the intended input domain and failure condition unique. A generic ``drop the item'' recommendation without the construct-validity diagnosis does not pass.}

\subsection{Instructional-action selection---representation: \texttt{EI-C09}}

\eabdesignfield{Evidence contract}{A calculus learner can reproduce the chain rule on practiced forms but fails novel compositions and can only recite a verbal template. Among four labeled interventions, a nested function-machine representation makes composition an object the learner can inspect, consistent with the process--object distinction~\citep{sfard1991dual}.}

\eabdesignfield{Discriminating design}{Twenty more exercises rehearse the procedural schema the learner already has; a limit proof introduces formalism before a structural concept image exists; cancellation notation can become another mnemonic. The case asks which representation repairs the diagnosed form of understanding, not which explanation of the chain rule is mathematically valid.}

\eabdesignfield{Matched verifier}{The response must diagnose procedural rather than structural understanding, select the labeled function-machine option, explain why drill-first fails, defer formal definition to a later stage, and connect the sequencing decision to an accepted learning theory.}

\subsection{Instructional-action selection---scaffolding: \texttt{EI-SCAFFOLD-FADE-01}}

\eabdesignfield{Evidence contract}{A tutoring coordinator must judge whether one tutor's support was too high or too low for each of two learners and prescribe concrete changes for their next sessions. The learners attempted the same six-step procedure; at every step, the task records whether each learner succeeded before tutor help and how much help followed. Avery fails the two hardest steps while help falls to 1 and 0; Blair succeeds on every step while the tutor continues at directive/full-help levels 3--4. Contingent scaffolding therefore prescribes more targeted help for Avery and less for Blair~\citep{wood1975assisted,wood1976role,vandepol2010scaffolding}.}

\eabdesignfield{Discriminating design}{A one-week check crosses the surface cue with the correct action: Avery scores 5/6 and Blair 3/6. Treating the lower score as a request for more help reverses both prescriptions. The trace supports a mechanism-level alternative: Avery encountered unsupported breakdowns, while Blair was denied independent practice and transfer of responsibility.}

\eabdesignfield{Matched verifier}{The response must compare unaided success to help level for each learner, prescribe opposite and correctly directed changes, and explicitly refute the score shortcut. Ability labels, the word ``scaffolding,'' or the two directions without their log evidence are insufficient.}

\endgroup
\endgroup

\section{\rev{Stage 2 Design Audit and Representative Cases}}
\label{app:stage2-design-audit}
\begingroup

Stage~2 contains 100 situated-tutoring tasks. The release preserves historical \texttt{S0-*} identifiers for backward compatibility; these are Stage~2 task IDs, not paper stage labels. Sixty-four tasks are fresh scenarios grounded in OER/MathDial misconception patterns; 36 transform MATH, MMLU-Pro, MIT OpenCourseWare, or topic-matched hard items. Every task fixes a fine-grained KC, known and unknown learner knowledge, predicted errors, and a changed-surface concept check. The auxiliary weak-model pair is audited separately: cold pre-test failure determines whether a fixed-model learning-gain result is reportable, not whether the task belongs in Stage~2, and the probe has zero headline weight.

\subsection{Episode construction and profile coverage}

We construct each episode from content outward. Annotators first resolve the source answer and label the fine-grained KC to be changed. They then specify the learner's correct foothold, missing link or generative misconception, and the errors that should follow from it; author a first-person opening that exposes this state without leaking the repair; and ground-truth a changed-surface concept check targeting the same KC. The source fixes what is true, while the authored learner--content contrast fixes what the tutor must notice and how the tutoring policy should differ.

The learner contract makes three policy-relevant dimensions explicit. Cognitive fields encode prior mastery, known and unknown knowledge, predicted errors, and explanation-load sensitivity; metacognitive fields encode forethought, monitoring, and reflection in Zimmerman's self-regulated-learning cycle; non-cognitive fields encode motivation, persistence, and achievement emotion. Big-Five and grit cues serve only as coarse behavioral priors, not as learner diagnoses~\citep{brown1978diagnostic,corbett1994knowledge,sweller1988cognitive,zimmerman2002self,ryan2000intrinsic,pekrun2006control,goldberg1990alternative,duckworth2007grit}. Explicit knowledge bounds address the tendency of LLM student simulators to drift into implausibly expert respondents~\citep{wu2025imperfection}.

For coverage, the tasks instantiate six authored composites: brittle experts (15), motivated novices (18), deep learners (19), strategic surface learners (18), anxious achievers (13), and disengaged/amotivated learners (17). These are controlled benchmark conditions, not six claimed natural kinds of student. They combine the three dimensions to create distinct tutoring pressures: for example, a curious novice can productively discover a missing link, whereas an anxious achiever may first need restored control without having the cognitive demand removed.

\begin{table}[H]
\caption{\rev{\textbf{Stage~2 learner-profile coverage.} This table expands the six controlled composites in Table~\ref{tab:stage2-learner-design} with task counts and full adaptation requirements. They are not asserted to be discrete natural learner types.}}
\label{tab:stage2-archetypes}
\centering
\scriptsize
\setlength{\tabcolsep}{3.2pt}
\renewcommand{\arraystretch}{1.06}
\begin{tabularx}{\textwidth}{>{\raggedright\arraybackslash}p{2.25cm}>{\raggedright\arraybackslash}p{3.9cm}X}
\toprule
\textbf{Composite (tasks)} & \textbf{Controlled learner pressure} & \textbf{Adaptation the tutor must demonstrate} \\
\midrule
Brittle expert (15) & High procedural fluency and confidence with one consequential conceptual hole. & Isolate and destabilize the faulty rule without reteaching mastered material or provoking face-saving resistance. \\
Motivated novice (18) & Limited prior knowledge, intrinsic curiosity, and willingness to reason. & Build from a correct fragment and let the learner perform the discriminating step rather than rewarding enthusiasm with an answer dump. \\
Deep learner (19) & Strong self-regulation and a preference for mechanisms over recipes. & Offer conceptual depth and transfer while avoiding remedial micro-steps the learner does not need. \\
Strategic surface learner (18) & Completion- or grade-oriented behavior and repeated pressure for the shortest answer. & Keep the next step low-cost but resist answer extraction; require explanation before closure. \\
Anxious achiever (13) & Substantial preparation coupled with error sensitivity and rapid loss of confidence. & Stabilize affect while preserving cognitive ownership; do not mistake capitulation for learning. \\
Disengaged/amotivated (17) & Low persistence, weak perceived value, and minimal replies under challenge. & Re-establish a reason and feasible entry point for engagement, then verify substance rather than equating verbosity with learning. \\
\bottomrule
\end{tabularx}
\end{table}

\subsection{Closed-loop simulation and matched judging}

\paragraph{Observable teaching loop.}
The target trace elicits current reasoning, diagnoses the active barrier, scaffolds the next attainable step, updates support after the learner responds, and then fades and verifies on a changed surface. Because the appropriate move is history-dependent, the same hint can be productive initially, neglectful after repeated failure, or a takeover after the learner can proceed independently. The tutor sees only the conversation; profile fields and evaluator state remain hidden.

\paragraph{Learner fidelity and state change.}
Each reply conditions on the profile, dialogue history, and an optional private self-regulation record. Learner generation uses \texttt{qwen3-max-2026-01-23} at \(T=0\). The implementation separates generation, fidelity screening, and learner-state updating into role-specific calls with distinct prompts and structured output contracts. The fidelity call checks profile and history consistency, cognitive bounds, metacognition and affect, over-compliance, and cross-turn drift; rejected replies regenerate. For every accepted reply, the state-update call compares the state before and after the reply, tracking KC repair, self-regulation, and affect or engagement. Application and explanation provide stronger evidence than assent or echoes, while revelation annotations discount apparent mastery after tutor-supplied reasoning. Only the fidelity-approved reply is visible to the tutor, keeping simulator fidelity distinct from measured learner change~\citep{chi2014icap,zimmerman2002self}.

\paragraph{Scoring realization.}
The three scopes in Table~\ref{tab:stage2-judging-dimensions} are matched but non-interchangeable. The tutor-turn scope evaluates one move in context, classifies its pedagogical strategy, and scores correctness, contingency, clarity and load, immediate affective response, elicited uptake, support, and information revelation. The trajectory scope combines structured aggregation over successive turn annotations with whole-dialogue assertions: it tests whether support escalates after failure and fades after success, whether the tutor changes representation or strategy when needed, whether agency returns to the learner, and whether diagnosis, repair, verification or transfer, and task-specific obligations are completed. Complementary evaluator views cover these semantic components. For replies that pass the fidelity check, the learner-change scope updates cognitive (50\%), metacognitive (25\%), and non-cognitive (25\%) state. Tutor-turn and trajectory scoring is applied post hoc and is distinct from the online simulator fidelity and state-update calls described above.
\begin{equation}
R_2=0.35R_{\mathrm{turn}}+0.16R_{\mathrm{traj\text{-}metric}}+0.24R_{\mathrm{traj\text{-}judge}}+0.25R_{\mathrm{learner}}.
\end{equation}
Ninety-eight of 100 tasks include learner outcome. Weights are renormalized over applicable and successfully measured components, and a session without an explicit completion signal receives a \(0.92\) multiplier. ICAP and revelation level calibrate learner-change credit: for example, mastery receives $0.20\!\times$ credit after a complete same-turn reveal, versus $0.70\!\times$ after partial scaffolding. Evaluator records store component availability and nominal weights from which the effective aggregation is derived.

\paragraph{Auxiliary same-KC transfer.}
In 86 tasks, two additional non-identical same-KC questions are reserved for a fixed weak model. The post-question is evaluated only when that model misses the cold pre-question. This diagnostic does not select tasks and has zero headline weight; it tests whether a trajectory helps one fixed proxy generalize beyond the tutored surface, not whether a human learned, and remains separate from the learner-change component above.

\subsection{Representative design cards}

The cards below were chosen because each learner preserves a correct fragment while embedding a consequential error. Answer checking can therefore reward the wrong mental model, whereas an effective tutor must elicit discriminating evidence, adapt to the learner, and return ownership before closure.

\newcommand{\eabtutorfield}[2]{%
  \par
  \noindent\colorbox{EABStageTwo!10}{\strut\textcolor{EABStageTwo}{\textbf{#1}}}\hspace{0.45em}#2\par
}

\begingroup
\small
\setlength{\parskip}{0pt}

\subsubsection{A defensible recommendation for a false reason (\texttt{S0-99})}

\eabtutorfield{Grounding and construction}{The supplied kidney-stone counts are deterministic: Treatment A has the higher cure rate within both stone-size groups, yet pooling reverses the rates to \(273/350=78.0\%\) for A and \(289/350=82.6\%\) for B. The learner's recommendation of A is defensible after conditioning on stone size, but their stated rule---a subgroup winner must also win in aggregate---is false. The case instantiates Simpson reversal~\citep{simpson1951interpretation}; a baseball transfer item changes the domain while preserving unequal mixture weights.}

\eabtutorfield{Discriminating policy}{Confirming 'choose A' rewards the wrong model; replacing it with 'choose B' rewards a different aggregate shortcut. The motivated novice is willing to calculate, so the tutor should first elicit both pooled rates without naming the reversal. If the learner averages subgroup percentages, support should make the unequal denominators inspectable; if the learner computes $78.0\%<82.6\%$ but jumps to ``B is better,'' support should elicit how case mix changes the comparison. Only after the learner explains that mechanism should support fade and the baseball transfer item appear.}

\eabtutorfield{Matched judge}{The selected Claude Opus~4.6 trace asks the learner to pool the counts, lets the reversal challenge the false rule, and then elicits the unequal-case-mix explanation. The selected GPT-5.5-Pro trace is factually correct and supportive, but its opening supplies both unequal weighting and the case-mix mechanism, so the learner need only accept them. Turn-level revelation, trajectory-level adaptation and completion, and learner-owned explanation therefore distinguish the traces even when their content is correct. Five task assertions bind diagnosis of the false subgroup-to-aggregate rule, both pooled rates before disclosure, the unequal-case-mix mechanism, curiosity-sensitive challenge, and a defensible comparison plus learner-owned transfer before closure.}

\subsubsection{A correct product with a false algebraic law (\texttt{S0-95})}

\eabtutorfield{Grounding and construction}{For \(A=\left[\begin{smallmatrix}1&2\\0&1\end{smallmatrix}\right]\) and \(B=\left[\begin{smallmatrix}1&0\\3&1\end{smallmatrix}\right]\), the learner correctly obtains \(AB=\left[\begin{smallmatrix}7&2\\3&1\end{smallmatrix}\right]\) but claims \(BA=AB\). Direct calculation gives \(BA=\left[\begin{smallmatrix}1&2\\3&7\end{smallmatrix}\right]\). Fresh matrices provide a deterministic transfer check in which both products again differ.}

\eabtutorfield{Discriminating policy}{The learner is procedurally fluent, impatient, and resistant to correction. Validating the message as a whole preserves imported scalar commutativity; announcing non-commutativity supplies an arbitrary-looking exception. The tutor should acknowledge only the verified \(AB\), request one low-cost entry of \(BA\), and let the learner's own procedure contradict the rule before explaining the row--column direction.}

\eabtutorfield{Matched judge}{Checks require atomic diagnosis of the false \(BA\) claim, learner-generated counterevidence before the rule is named, a mechanism tied to reversed row--column roles, face-preserving but non-capitulating feedback, and a fresh product pair before closure. This separates content completion from a policy that can revise a brittle expert's belief.}

\subsubsection{Fluent substitution with meaningless bounds (\texttt{S0-110})}

\eabtutorfield{Grounding and construction}{For \(\int_0^2 x(x^2+1)^3\,dx\), the learner correctly chooses \(u=x^2+1\) and obtains \(u^4/8\), then inserts the original \(x\)-bounds into a function of \(u\) and reports 2. Converting \(x=0,2\) to \(u=1,5\) gives \((5^4-1^4)/8=78\). A new integral changes the substitution and bounds while preserving the representational obligation.}

\eabtutorfield{Discriminating policy}{Saying 'use 1 and 5' repairs one answer while leaving the mental model untouched. This strategic learner explicitly requests the shortcut, so the tutor must ask what variable each endpoint denotes and require one coherent route: convert the bounds to \(u\), or back-substitute the antiderivative into \(x\). The desired restraint is pedagogical, not conversational austerity.}

\eabtutorfield{Matched judge}{Turn checks penalize revealing the converted bounds or 78 before learner reasoning. Trajectory checks require the learner to identify the variable--bound mismatch, repair the original integral, and independently preserve bound meaning in the transfer problem. Task-specific assertions prevent generic praise for otherwise fluent substitution mechanics.}

\subsubsection{Inverse probability under anxiety (\texttt{S0-54})}

\eabtutorfield{Grounding and construction}{With \(1\%\) prevalence, \(99\%\) sensitivity, and a \(5\%\) false-positive rate, a population of 10,000 yields 99 true positives and 495 false positives, so the posterior is \(99/(99+495)=1/6\), not \(0.99\). The learner correctly recalls the sensitivity but treats \(P(+\mid D)\) as \(P(D\mid+)\). Natural frequencies make the two reference populations inspectable, and a commuter/rain transfer item changes the surface context.}

\eabtutorfield{Discriminating policy}{The learner is conscientious but anxious and will abandon even a correct fragment when the tutor sounds skeptical. Blunt correction or formula dumping can therefore yield compliance without reconstructing the conditional direction. The policy should stabilize affect, preserve the correctly identified sensitivity, then let the learner count true and false positives and form the denominator.}

\eabtutorfield{Matched judge}{Table~\ref{tab:stage2-case-judge-trace} summarizes an evaluated trace in which the layers separate cleanly. The turn judge credits the affect-first natural-frequency prompt, but marks a later correct explanation as substantial revelation because it supplies the key interpretation that most positives are false. The trajectory layer credits the state-contingent sequence from reassurance under anxiety to lighter support after successful calculation and the completed conceptual repair, while flagging weak agency handoff and the absence of a changed-surface transfer inside the tutor dialogue. The learner-change layer credits the learner's independent \(1/6\) calculation, base-rate explanation, and shift from anxious confirmation-seeking to explanatory engagement rather than the tutor's words. The reserved commuter/rain item provides an auxiliary same-KC transfer check; it is not counted as a transfer move made by the tutor inside this dialogue.}

\subsubsection{Correct direction, unsupported magnitude (\texttt{S0-46})}

\eabtutorfield{Grounding and construction}{The learner correctly predicts that a \(20\%\) price increase lowers quantity but silently assumes a one-for-one \(20\%\) quantity decline, ignoring the supplied point elasticity \(E=-0.5\). Under the task's explicitly local-linear convention, \(\Delta Q/Q\approx E\,\Delta P/P\) gives an estimated \(10\%\) decline, approximately 90 units, and approximately \(\$1{,}080\) revenue at \(\$12\). Point elasticity alone does not make these exact finite-change consequences without a demand curve or elasticity path.}

\eabtutorfield{Discriminating policy}{The deep learner already owns the direction and revenue identity; the missing link is magnitude. A narrow question---where did the \(20\%\) quantity change come from?---lets the learner locate the skipped representation, compare price and quantity effects, and explain why the price effect dominates locally under inelastic demand. Giving the slogan 'inelastic means revenue rises' would bypass that repair.}

\eabtutorfield{Matched judge}{The judge checks diagnosis of direction-versus-magnitude confusion, Socratic use of the elasticity relation, reconstruction of the revenue mechanism, and a transfer case in which elastic demand reverses the revenue conclusion. Turn and trajectory views distinguish learner-owned causal repair from a first-turn dump of formula, quantity, revenue, and rule.}

\endgroup
\endgroup

\section{\rev{Stage 3 Design Audit and Representative Cases}}
\label{app:stage3-design-audit}
\begingroup

Stage~3 contains 137 end-to-end LMS teaching workflows, each delegating a bounded instructor objective to the agent. The five cards below complement the main-text \texttt{MM-04} case by sampling additional evidence--decision--action dependencies rather than merely different tools. Each card is written as a miniature measurement contract: the recoverable evidence, the decision it warrants, the durable course outcome, the strongest plausible wrong branch, and the verifier evidence that separates them. All records are synthetic, and all case IDs point to released task definitions.

\subsection{Construction and release audit}

We applied an eight-class construction audit to the release-candidate pool. For Stage~3, construction oracles were re-executed from seeded initial states across this pool; automated adversarial review and representative-trajectory inspection separately assessed semantic alignment with each task's educational rationale. Defects found during construction were repaired before model comparison, with review organized around the insight--evidence--trace--verifier chain rather than a desired ranking.

\subsection{Scoring realization}

Each workflow declares only the verifier components its contract makes observable. All 137 tasks include environment and semantic checks; subsets additionally declare goal state, process, artifact quality, or a response anchor (Table~\ref{tab:verifier-coverage}). Programmatic checks evaluate persistent state and required ordering, while rubric-based checks evaluate semantic and artifact quality. Artifact quality enters multiplicatively rather than additively and cannot rescue a zero required state component, while completed non-critical portions retain diagnostic partial credit. Declared penalties cover visible backend-language leakage and tool-error-limit termination; blind-grading violations are hard failures. The task manifests record which checks apply to each workflow.

\newcommand{\eabworkflowfield}[2]{%
  \par
  \noindent\colorbox{EABStageThree!10}{\strut\textcolor{EABStageThree}{\textbf{#1}}}\hspace{0.45em}#2\par
}

\begingroup
\small
\setlength{\parskip}{0pt}

\subsection{Evidence-conditioned assessment: \texttt{AD-01}}

\eabworkflowfield{Evidence contract}{A Fall~2023 PHYS midterm supplies KC-level performance for constructing a 12-question final. The weakest targets are \texttt{torque\_equilibrium} (\(6.7\%\)), \texttt{circular\_motion} (\(13.3\%\)), \texttt{projectile\_time\_of\_flight} (\(18.3\%\)), and \texttt{free\_body\_diagram} (\(22.5\%\)). The new assessment must emphasize evidenced weaknesses while retaining broad cognitive-level coverage across at least three Bloom levels and limiting the strongest KC, \texttt{newton\_second\_law\_basic} (\(65.0\%\)), to at most one item.}

\eabworkflowfield{Discriminating design}{A generic 12-item exam can look balanced while ignoring the historical evidence; a lowest-rate-only policy can overfit one KC while violating cognitive-level coverage. The task therefore separates surface completeness from an assessment blueprint justified by both student data and construct coverage.}

\eabworkflowfield{Required trace and verifier}{The agent must retrieve the exact historical assessment before creating the final. Environment predicates require the named PHYS101 exam with all 12 questions and both weak-KC emphasis and broad course coverage. Scenario-specific semantic checks inspect the non-uniform weak-versus-strong allocation, multi-level Bloom coverage, and evidence-based rationale; the process constraint requires the F23 record before creation; and the quiz-quality rubric checks keyed-answer correctness, construct alignment, distractors, calibration, clarity, and cognitive diversity. Together they distinguish an uploaded exam from one whose blueprint and items are warranted by the records.}

\subsection{Diagnosis-conditioned feedback: \texttt{CE-01}}

\eabworkflowfield{Evidence contract}{Three learners submit factorial implementations. Two solutions pass the executable tests; the third fails at \(n=0\) because its base case is \(n=1\), causing unbounded recursion. Historical course evidence shows that base-case necessity is a class-level weakness (\(36/120=30\%\)). The correct intervention combines individual grading, private failure-specific feedback for the affected learner, and a privacy-preserving class reminder about the shared KC.}

\eabworkflowfield{Discriminating design}{Giving all three learners the same generic recursion advice ignores executable evidence; privately messaging every learner leaks an inference that applies to only one submission; posting the failing learner's identity to the class violates the communication contract. The task makes diagnosis, audience, and action scope jointly necessary.}

\eabworkflowfield{Required trace and verifier}{The task provides three implementations and explicit tests. State checks verify the resulting grade pattern and recipient sets; semantic checks require exact test outcomes, the concrete base-case failure, and a class message that discusses the shared KC without exposing the learner. A separate process check requires retrieving the historical KC evidence before remediation messages are sent.}

\subsection{Persistent-gap remediation: \texttt{LW-02}}

\eabworkflowfield{Evidence contract}{A 120-student Fall~2023 baseline and six-student Fall~2024 midterm identify two gaps that persist across cohorts: law of total probability, at \(48/120=40.0\%\) and \(1/6=16.7\%\), and enumerating all mutually exclusive event cases before summing, at \(52/120=43.3\%\) and \(2/6=33.3\%\). By contrast, conditional-probability definition was \(87/120=72.5\%\) historically, so its small-cohort decline is not evidence of a persistent weakness.}

\eabworkflowfield{Discriminating design}{A current-cohort-only analysis promotes sampling noise to curriculum policy. The environment also contains an editable regression deck and permits creating a new file, so topical prose in the wrong object is an easy operational success. The valid intervention must preserve the cross-cohort diagnosis in the assigned conditional-probability deck and a matched quiz.}

\eabworkflowfield{Required trace and verifier}{Six process constraints require both exact assessments before computation or writes, deck inspection before editing, at least two existing-slide updates, and the slide and quiz actions before announcement. Eight environment predicates inspect the named deck, slide content, aligned three-question quiz, and evidence-bearing announcement; the generic goal state independently checks quiz and announcement existence. Five semantic assertions bind all outputs to the two persistent KCs, and a slide-quality rubric checks conceptual clarity, worked examples, KC targeting, cognitive load, and engagement.}

\subsection{Audience and privacy control: \texttt{COMM-03}}

\eabworkflowfield{Evidence contract}{The instructor's risk rule is grade below 50 or attendance below 40\%. The records identify \texttt{stu\_di} (25.5, 22\%), \texttt{stu\_mn} (41.2, 38\%), and \texttt{stu\_be} (48.0, 85\%) as the exact recipients; \texttt{stu\_ss} is a designed near miss at 50.4 and 72\%. The required communication separates an anonymous class-level pattern, individual messages containing only each learner's own evidence, and an advisor summary containing the full authorized detail.}

\eabworkflowfield{Discriminating design}{Using a non-strict \(\leq 50\) cutoff instead of the declared \(<50\) rule incorrectly contacts the near-boundary learner; one templated message cannot satisfy three audience-specific privacy obligations; a class post with names or individual records is informative but invalid. The case tests whether strict recipient selection survives disclosure boundaries.}

\eabworkflowfield{Required trace and verifier}{The trace must retrieve the governing records before messaging. Environment checks bind exact recipient sets and absence of false-positive contacts; process checks enforce evidence-before-communication; semantic checks verify the risk rationale, learner-specific evidence, class anonymity, and advisor-only aggregation.}

\subsection{Construct-preserving revision: \texttt{EXAM-REV-03}}

\eabworkflowfield{Evidence contract}{Five specified calculus items must receive new parameters while preserving their KCs and solution techniques: quotient rule, two related-rates settings, tangent line, and an exponential-versus-polynomial limit. The revised worked answers must use the new parameters; in particular, the final exponential must still dominate the polynomial so that the limit remains zero.}

\eabworkflowfield{Discriminating design}{Grammatical paraphrase can silently change the construct, and recomputing from the original parameters produces a polished but false answer key. The case therefore treats item identity, KC invariance, parameter change, and mathematical consistency as a linked revision obligation.}

\eabworkflowfield{Required trace and verifier}{The reference trajectory inspects the source exam, runs a symbolic/numeric sanity check, creates the exact named document, and then announces it. Scored process constraints require source inspection before calculation or document creation and the document before the announcement. Environment predicates require the five IDs, verified keys, KC-preservation language, and announcement; artifact-grounded semantic checks recompute whether each answer follows the changed parameters and enforce the zero-limit invariant. The executable calculation is demonstrated by the reference trajectory but is not itself a separate required-call constraint, so final mathematical consistency is judged from the artifact rather than trusted from tool use alone.}

\endgroup
\endgroup

\section{\rev{Evaluation and Aggregation Details}}
\label{app:evaluation-details}
\begingroup

The benchmark release fixes task definitions, source packs, environment and learner-profile fields, and stage-specific verifier contracts. Continuous component rewards expose partial progress and localize the unsatisfied part of a task contract. Because the stages observe different evidence---a static teacher decision, a tutor--learner trajectory, and a persistent workflow state---their scores are complementary diagnostic surfaces rather than interchangeable scales.

The scoring stack follows the evidence available at each stage. Stage~1 requires a warranted decision and mechanism-sensitive justification, with a programmatic response anchor when the decision admits one. Stage~2 combines individual tutor moves, the policy expressed across the full episode, and learner-produced change discounted when the tutor has already revealed the reasoning. Stage~3 conditionally composes semantic, persistent-state, process, and artifact checks according to the task manifest. Stage-specific scoring realizations and applicability rules are described in Appendices~\ref{app:stage1-design-audit}--\ref{app:stage3-design-audit}.

\begin{table}[H]
\caption{\rev{\textbf{Verifier coverage in the official 354-task release.} Counts report declared scoring components; the Stage~2 transfer probe is logged but has zero leaderboard weight.}}
\label{tab:verifier-coverage}
\centering
\scriptsize
\setlength{\tabcolsep}{3pt}
\renewcommand{\arraystretch}{1.08}
\begin{tabularx}{\linewidth}{@{}>{\raggedright\arraybackslash}p{2.15cm}>{\centering\arraybackslash}p{0.55cm}X@{}}
\toprule
\textbf{Surface} & \textbf{\(N\)} & \textbf{Verifier coverage} \\
\midrule
Stage~1 judgment & 117 & Semantic assertions 117; programmatic response anchors 100. \\
Stage~2 tutoring & 100 & Turn policy 100; trajectory 100; learner outcome 98; auxiliary transfer probe 86. \\
Stage~3 workflows & 137 & Environment 137; semantic 137; process 123; goal state 87; artifact quality 83; response anchors 2. \\
\bottomrule
\end{tabularx}
\end{table}

The reported board summarizes \eabCompleteOverlays{} model overlays against the \eabTotalTasks{}-task manifest. Within each stage, we average the recorded applicable task rewards and compute
\[
\bar R_{\mathrm{bal}}=\frac{\bar R_1+\bar R_2+\bar R_3}{3},
\]
where each stage score is the mean reward within that stage, so unequal task counts do not determine the headline ranking.

The public task snapshot is pinned to Hugging Face commit \href{https://huggingface.co/datasets/CinderD/TeachArena/tree/cbd99fcca76b076352388d01be10056a5ee47eb4}{\texttt{cbd99fcca76b}}. For backward compatibility, its subset names preserve the original release numbering: paper Stage~1 corresponds to \texttt{stage1\_pedagogical\_judgment}, paper Stage~2 to \texttt{stage0\_situated\_tutoring}, and paper Stage~3 to \texttt{stage2\_teaching\_workflows}. Task IDs such as \texttt{S0-54} are therefore historical identifiers rather than paper-stage labels.

\endgroup

\section{Additional Aggregate Analyses}
\label{app:diagnostic-results}

\paragraph{Analysis protocol.}
We compute the subgroup summaries in Section~\ref{sec:results} from a frozen task-to-condition analysis map and the 17 reported systems, averaging scores within each condition. Confidence intervals use 10,000 stratified task-cluster bootstrap replicates (seed 20260731), resampling tasks within the focal and comparison strata so that all model evaluations of one task remain together. We examine two theoretically motivated contrasts: shortcut seeking as a stressor for learner agency and long workflows as a stressor for cross-interface composition. The intervals quantify task-sampling uncertainty for these descriptive contrasts, not causal effects.

\begin{table}[t]
\caption{\textbf{Complete subgroup context for the two aggregate analyses.} Means are computed across the 17 reported systems. Rows are ordered by mean within each panel; the main text discusses learner agency and multi-interface composition, while this table reports every profile and workflow family.}
\label{tab:subgroup-results}
\centering
\scriptsize
\setlength{\tabcolsep}{4pt}
\renewcommand{\arraystretch}{0.96}
\begin{tabularx}{0.96\textwidth}{@{}>{\raggedright\arraybackslash}Xrr@{}}
\toprule
\textbf{Controlled condition} & \textbf{Tasks} & \textbf{Mean} \\
\midrule
\multicolumn{3}{@{}l}{\textbf{Stage 2 learner profile}} \\
Strategic surface learner & 18 & 0.702 \\
Disengaged/amotivated & 17 & 0.736 \\
Anxious achiever & 13 & 0.753 \\
Deep learner & 19 & 0.760 \\
Brittle expert & 15 & 0.764 \\
Motivated novice & 18 & 0.768 \\
\midrule
\multicolumn{3}{@{}l}{\textbf{Stage 3 workflow family}} \\
Long-chain workflows & 6 & 0.321 \\
Multimodal/material generation & 8 & 0.472 \\
Communication & 20 & 0.519 \\
Feedback/grading & 17 & 0.526 \\
Content/curriculum/assessment & 15 & 0.548 \\
Adaptive review/mixed workflows & 22 & 0.555 \\
Learning analytics & 12 & 0.566 \\
Documents/pages/reports/worksheets & 10 & 0.584 \\
Gradebook/goal state/spreadsheets & 13 & 0.623 \\
Advising/prerequisites/differentiation & 14 & 0.641 \\
\bottomrule
\end{tabularx}
\end{table}

\paragraph{Stage 2 learner profiles.}
The strategic surface learner is the lowest-scoring of the six controlled profiles for 15 of 17 models. Averaged across evaluated cases, its score is 0.702, compared with 0.757 across the other profiles. The resulting 0.055 gap has a stratified task-cluster bootstrap 95\% confidence interval of 0.033--0.078. This profile is deliberately interactive rather than merely unresponsive: the learner participates but requests shortcuts or answer confirmation, requiring the tutor to preserve productive struggle and elicit learner-owned reasoning.
The two exceptions are Claude Opus~4.6 and Kimi~K2.6, for which the disengaged/amotivated profile is lowest.

\paragraph{Stage 3 workflow composition.}
The six long-chain workflows average 0.321, compared with 0.559 across the other workflow families, and are the lowest-scoring family for 15 of 17 models. The 0.238 gap has a stratified task-cluster bootstrap 95\% confidence interval of 0.170--0.302. Long-chain tasks jointly require evidence retrieval, pedagogical reasoning, persistent-state or artifact operations, and communication. The association is consistent with composition difficulty, but does not identify chain length as its causal source.
The two exceptions are Claude Opus~4.6, whose lowest family is multimodal/material generation, and GPT-5.5-Pro, whose lowest family is learning analytics.

\section*{NeurIPS Paper Checklist}

\begin{enumerate}

\item {\bf Claims}
    \item[] Question: Do the main claims made in the abstract and introduction accurately reflect the paper's contributions and scope?
    \item[] Answer: \answerYes{}.
    \item[] Justification: The abstract and introduction state the \eabTotalTasks{}-task benchmark scope, evaluation protocol, public release, and main empirical finding. The limitations and data-availability section explicitly separates benchmark claims from deployment certification.

\item {\bf Limitations}
    \item[] Question: Does the paper discuss the limitations of the work performed by the authors?
    \item[] Answer: \answerYes{}.
    \item[] Justification: Section~7 discusses the limitations of the work, including the fact that \textsc{\systemName{}} is a simulation benchmark rather than a deployment certification, the limits of mock learners and judge models, the auxiliary nature of transfer probes, source/provenance constraints, and the need for studies with real students and instructors.

\item {\bf Theory assumptions and proofs}
    \item[] Question: For each theoretical result, does the paper provide the full set of assumptions and a complete (and correct) proof?
    \item[] Answer: \answerNA{}.
    \item[] Justification: The paper is an empirical benchmark paper and does not introduce formal theoretical results or proofs.

    \item {\bf Experimental result reproducibility}
    \item[] Question: Does the paper fully disclose all the information needed to reproduce the main experimental results of the paper to the extent that it affects the main claims and/or conclusions of the paper (regardless of whether the code and data are provided or not)?
    \item[] Answer: \answerYes{}.
    \item[] Justification: Sections~3 and 4 describe the benchmark construction, simulator modes, evaluation protocol, reward components, official thresholds, task-family pass criteria, and model evaluation setup. The appendices further document task provenance, verifier design, reward rules, audit protocol, and diagnostic cases.

\item {\bf Open access to data and code}
    \item[] Question: Does the paper provide open access to the data and code, with sufficient instructions to faithfully reproduce the main experimental results, as described in supplemental material?
    \item[] Answer: \answerYes{}.
    \item[] Justification: The benchmark is publicly released at \url{https://huggingface.co/datasets/CinderD/TeachArena} with task definitions, metadata, evaluator code, evaluation criteria, and a synthetic LMS environment containing no real learner records.

\item {\bf Experimental setting/details}
    \item[] Question: Does the paper specify all the training and test details (e.g., data splits, hyperparameters, how they were chosen, type of optimizer) necessary to understand the results?
    \item[] Answer: \answerYes{}.
    \item[] Justification: Section~4 specifies the evaluation protocol, official task thresholds, stage-wise scoring, simulator setup, source-safety filtering, complete-overlay reporting, and the distinction between dense rewards and official pass rates. Additional details are provided in the appendices, including task families, process constraints, reward components, and audit procedures.

\item {\bf Experiment statistical significance}
    \item[] Question: Does the paper report error bars suitably and correctly defined or other appropriate information about the statistical significance of the experiments?
    \item[] Answer: \answerNo{}.
    \item[] Justification: Appendix~F reports 95\% stratified task-cluster bootstrap confidence intervals for the two subgroup contrasts. The current snapshot reports deterministic leaderboard runs and diagnostic overlays rather than multi-seed confidence intervals. We disclose this limitation and treat multi-seed variance measurement as future work.

\item {\bf Experiments compute resources}
    \item[] Question: For each experiment, does the paper provide sufficient information on the computer resources (type of compute workers, memory, time of execution) needed to reproduce the experiments?
    \item[] Answer: \answerNo{}.
    \item[] Justification: The paper describes the evaluation protocol and model coverage, but does not yet provide a full hardware, memory, wall-clock, and cost accounting for every run.
    
\item {\bf Code of ethics}
    \item[] Question: Does the research conducted in the paper conform, in every respect, with the NeurIPS Code of Ethics \url{https://neurips.cc/public/EthicsGuidelines}?
    \item[] Answer: \answerYes{}.
    \item[] Justification: The work is an evaluation benchmark for educational agents, does not deploy agents to learners, and includes release filtering and limitations intended to reduce misleading or unsafe claims.

\item {\bf Broader impacts}
    \item[] Question: Does the paper discuss both potential positive societal impacts and negative societal impacts of the work performed?
    \item[] Answer: \answerYes{}.
    \item[] Justification: The paper discusses positive impacts, such as improving transparency and diagnostic evaluation for tutor agents, as well as negative impacts, such as overclaiming deployment readiness, relying on imperfect judge models, and misusing benchmark scores as certification for real educational deployment.
    
\item {\bf Safeguards}
    \item[] Question: Does the paper describe safeguards that have been put in place for responsible release of data or models that have a high risk for misuse (e.g., pre-trained language models, image generators, or scraped datasets)?
    \item[] Answer: \answerYes{}.
    \item[] Justification: The benchmark release process includes source-safe filtering, provenance metadata, exclusion of restricted raw source material, task-contract validation, and framing of the artifact as an evaluation benchmark rather than a deployed tutoring system. These safeguards reduce risks from misleading claims or inappropriate reuse of restricted educational materials.

\item {\bf Licenses for existing assets}
    \item[] Question: Are the creators or original owners of assets (e.g., code, data, models), used in the paper, properly credited and are the license and terms of use explicitly mentioned and properly respected?
    \item[] Answer: \answerYes{}.
    \item[] Justification: The source-grounding section and provenance appendices describe the public educational sources, license filtering, and derivative task-construction process used for releasable items.

\item {\bf New assets}
    \item[] Question: Are new assets introduced in the paper well documented and is the documentation provided alongside the assets?
    \item[] Answer: \answerYes{}.
    \item[] Justification: The benchmark introduces new task definitions, rubrics, simulator metadata, release filters, and audit artifacts, with documentation in the main paper and appendices.

\item {\bf Crowdsourcing and research with human subjects}
    \item[] Question: For crowdsourcing experiments and research with human subjects, does the paper include the full text of instructions given to participants and screenshots, if applicable, as well as details about compensation (if any)? 
    \item[] Answer: \answerNA{}.
    \item[] Justification: The reported benchmark construction and model evaluation do not involve crowdsourcing experiments or human-subject experiments with recruited participants.

\item {\bf Institutional review board (IRB) approvals or equivalent for research with human subjects}
    \item[] Question: Does the paper describe potential risks incurred by study participants, whether such risks were disclosed to the subjects, and whether Institutional Review Board (IRB) approvals (or an equivalent approval/review based on the requirements of your country or institution) were obtained?
    \item[] Answer: \answerNA{}.
    \item[] Justification: The paper does not report research with human subjects, so IRB approval is not applicable to the reported experiments.

\item {\bf Declaration of LLM usage}
    \item[] Question: Does the paper describe the usage of LLMs if it is an important, original, or non-standard component of the core methods in this research? Note that if the LLM is used only for writing, editing, or formatting purposes and does \emph{not} impact the core methodology, scientific rigor, or originality of the research, declaration is not required.
    \item[] Answer: \answerYes{}.
    \item[] Justification: LLMs are central to the benchmark setting and are also used in construction, audit, natural-language judging, and auxiliary transfer probes; these roles are described in the methods and evaluation sections.

\end{enumerate}

\end{document}